\documentclass[sigconf]{acmart}

\usepackage{booktabs} 
\usepackage{bm}
\usepackage{multirow,array}
\usepackage[linesnumbered,ruled,vlined]{algorithm2e}
\usepackage{balance}
\graphicspath{{figs/}}

\copyrightyear{2018} 
\acmYear{2018} 
\setcopyright{acmcopyright}
\acmConference[KDD '18]{The 24th ACM SIGKDD International Conference on Knowledge Discovery \& Data Mining}{August 19--23, 2018}{London, United Kingdom}
\acmBooktitle{KDD '18: The 24th ACM SIGKDD International Conference on Knowledge Discovery \& Data Mining, August 19--23, 2018, London, United Kingdom}
\acmPrice{15.00}
\acmDOI{10.1145/3219819.3219895}
\acmISBN{978-1-4503-5552-0/18/08}

\begin{document}
\title[Deep Sequence Learning with Auxiliary Information for Traffic Prediction]{Deep Sequence Learning with Auxiliary Information \\ for Traffic Prediction}
\titlenote{Binbing Liao and Jingqing Zhang contributed equally to this article. Fei Wu, Chao Wu and Yike Guo are the corresponding authors.This work was done while Binbing Liao was a research intern at Baidu.}



\author{Binbing Liao}
\affiliation{\institution{College of Computer Science and Technology, Zhejiang University}}
\email{bbliao@zju.edu.cn}

\author{Jingqing Zhang}
\affiliation{\institution{Data Science Institute, Imperial College London}}
\email{jingqing.zhang15@imperial.ac.uk}

\author{Chao Wu}
\affiliation{\institution{School of Public Affairs, Zhejiang University}}
\email{chao.wu@zju.edu.cn}

\author{Douglas McIlwraith}
\affiliation{\institution{Data Science Institute, Imperial College London}}
\email{dm05@imperial.ac.uk}

\author{Tong Chen}
\affiliation{\institution{College of Computer Science and Technology, Zhejiang University}}
\email{ckctonychen@zju.edu.cn}

\author{Shengwen Yang}
\affiliation{\institution{Baidu Inc.}}
\email{yangshengwen@baidu.com}

\author{Yike Guo}
\affiliation{\institution{Data Science Institute, Imperial College London}}
\email{y.guo@imperial.ac.uk}

\author{Fei Wu}
\affiliation{\institution{College of Computer Science and Technology, Zhejiang University}}
\email{wufei@zju.edu.cn}

\renewcommand{\shortauthors}{Binbing Liao, Jingqing Zhang et al.}

\begin{abstract}
Predicting traffic conditions from online route queries is a challenging task as there are many complicated interactions over the roads and crowds involved. In this paper, we intend to improve traffic prediction by appropriate integration of three kinds of implicit but essential factors encoded in auxiliary information. We do this within an encoder-decoder sequence learning framework that integrates the following data: 1) offline geographical and social attributes. For example, the geographical structure of roads or public social events such as national celebrations; 2) road intersection information. In general, traffic congestion occurs at major junctions; 3) online crowd queries. For example, when many online queries issued for the same destination due to a public performance, the traffic around the destination will potentially become heavier at this location after a while. Qualitative and quantitative experiments on a real-world dataset from Baidu have demonstrated the effectiveness of our framework.
\end{abstract}

%
%


\maketitle

\section{Introduction}
Traffic prediction is an important part of intelligent transportation systems (ITS) and is crucial to many applications including traffic network planning, route guidance, and congestion avoidance \cite{zhang2011data}. For large cities such as Beijing, such prediction is crucial but challenging to perform. This is due to the dynamic and complex traffic environment in large cities and the limited potential for new roads: this places an important emphasis on network management. Such management has a wide-reaching impact, not just due to the large population involved, but also because it supports decision making in various other applications e.g. the optimization of pollution is relevant to traffic. In this paper, we argue that previous models have failed to effectively include several important factors for prediction. We outline these as follows, before introducing our system which utilises these and demonstrates a positive impact on our application.

\textbf{Offline geographical and social factors}. The geographical structure of roads has an impact on traffic dynamics. For example, the traffic on a main road would be different from that of a lane and in general traffic congestion occurs more often at a major junction. Furthermore, social temporal factors such as holidays and the weekend have an influence on traffic. These characteristics serve to highlight the difficulty of traffic prediction. 

\begin{figure}[htbp]
\centering
\includegraphics[width=0.45\textwidth]{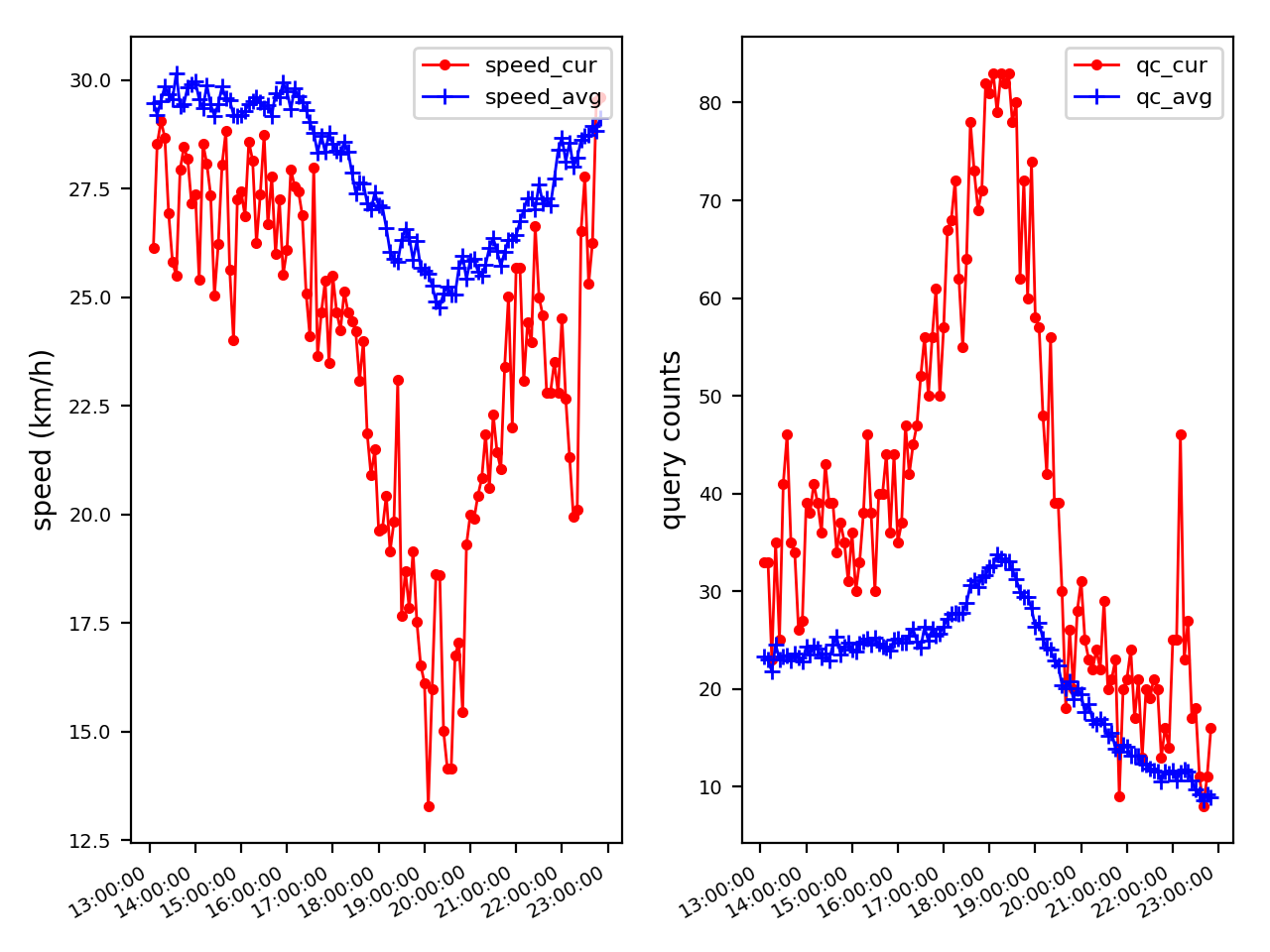}
\caption{The traffic speed (left) and online query counts (right) around the Beijing Capital Gym on April 8, 2017. The red line denotes the unusual traffic speed (query counts) while the blue line indicates the usual traffic speed (query counts). At 19:00 PM, there was the Fish Leong Concert in the Capital Gym.}
\label{fig:querytraffic}
\end{figure}

\textbf{Online potential influence}. Widely used mobile technology applications such as Baidu Map and Google Map provide a rich source of data for transportation analysis and forecasting. Figure \ref{fig:querytraffic} shows the average traffic speed and crowd query counts around Capital Gym, Beijing on April 8, 2017. The query counts at time $t$ are calculated by accumulating the queries whose destinations are around Capital Gym and their estimated arrival time is $t$. We can clearly observe that the current query counts (in red) are much more than usual query counts (in blue) at 18:00, which leads to a sudden drop of the traffic speed. Note that the query is long-term (the average travel time is 46 minutes shown in Table \ref{tab:querystats}) foreseeable, which would provide an early warning of traffic jams in ITS. More interestingly, the emergence of map queries from the crowd indicates that there is an event here, namely ``Fish Leong Concert". Online map queries from the crowd, which are innately related to the future states of road networks, can potentially be used to predict traffic dynamics, making the integration of multi-modal data an interesting yet challenging problem. 

\textbf{Limited dataset}. Due to the accessibility of traffic data, previous research on traffic prediction usually use limited datasets for experimentation. There are relatively few publicly available large-scale traffic prediction datasets for researchers to compare their models and propose new models.

To improve the state of the art in traffic prediction, we release a large-scale traffic dataset from Baidu Map, the Q-Traffic dataset, which provides various offline and online auxiliary information along with traffic speed data. There are three kinds of auxiliary domains in the Q-Traffic dataset: 1) Offline geographical and social attributes which include public holidays, peak-hour, speed etc; 2) the road intersection information such as local road network and junctions; and 3) online crowd queries which record map search queries from users. 

Table \ref{tab:linkproperties} shows that the offline geographical and social attributes from the Q-traffic dataset include a large number of categorical features, making the input features space sparse, i.e., the field \textit{speedclass} contains 8 class of speed limit. Learning and exploiting the sparse feature through a wide feature transformation is effective and interpretable, but requires a large degree of feature engineering effort. Conversely, deep neural networks can generalize better through low-dimensional dense embeddings. Motivated by \cite{cheng2016wide}, a wide transformation is utilised to learn the interactions from the sparse geographical and social attributes while an encoder-decoder deep network is used to decode the traffic condition given the combination of the traffic encoding and transformed features. 

Furthermore, due to the spatial dependencies in the road network, it is natural to utilise the graph convolutional neural networks \cite{niepert2016learning} to embed the traffic conditions induced by neighbouring road segments. Specifically, given a central road segment, neighbouring road segments are selected based on the PageRank score \cite{page1999pagerank} at first, which measures local spatial importance. The combination of graph CNN and the encoder-decoder deep network can integrate spatial patterns and historical traffic sequence.

Online map queries from the crowd, which are innately related to the future states of road networks, would potentially influence the traffic condition. For example, assume that we had the historical traffic data of road segments around the Capital Gym by 17:00 PM, being aware of many people would arrive at the Beijing Capital Gym around 18:00 PM, would effectively boost the performance of traffic prediction at 18:00 PM. We quantify the potential impact that online crowd queries have on the road segments, embed the impact with an encoder, and assemble the traffic and query impact with a deep fusion.

In this paper, we effectively utilise three kinds of auxiliary information in an encoder-decoder sequence to sequence (Seq2Seq) \cite{cho2014learning,sutskever2014sequence} learning manner as follows: a wide linear model is used to encode the interactions among geographical and social attributes, a graph convolution neural network is used to learn the spatial correlation of road segments, and the query impact is quantified and encoded to learn the potential influence of online crowd queries. Therefore a hybrid model based on deep sequence learning with auxiliary information for traffic prediction is proposed. In this model, offline geographical and social attributes, spatial dependencies and online crowd queries are integrated. The contribution of this paper can be summarised as follows:
\begin{itemize}
\item We release a large-scale traffic prediction dataset with offline and online auxiliary information including map crowd search queries, road intersections and geographical and social attributes.
\item We integrate the sequence to sequence deep neural networks with geographical and social attributes via a wide and deep manner.
\item To incorporate the spatial dependencies within local road network, we utilise the graph convolution neural network to embed the traffic speed of neighbouring road segments.
\item We quantify the potential influence and devise a query impact algorithm to calculate the impact that online crowd queries have on the road segments. 
\item We propose a hybrid Seq2Seq model which incorporates the offline geographical and social attributes, spatial dependencies and online crowd queries with a deep fusion.
\end{itemize}

The rest of this paper is organised as follows. We first introduce the released dataset in Sec.\ref{q-traffic}. Following that, we propose a series of methods for traffic prediction on the Q-Traffic dataset in Sec.\ref{methods}. Sec.\ref{experiments} presents qualitative and quantitative results of different methods. Sec.\ref{relatedwork} presents the related work of traffic prediction. Finally, we conclude the paper in Sec.\ref{conclusions}.

\section{Q-Traffic Dataset}
\label{q-traffic}
In this section, we first introduce a large-scale traffic prediction dataset --- Q-Traffic dataset \footnote{The code and dataset are available at https://github.com/JingqingZ/BaiduTraffic}, which consists of three sub-datasets: query sub-dataset, traffic speed sub-dataset and road network sub-dataset. We compare our released Q-Traffic dataset with different datasets used for traffic prediction.

\begin{table*}[htbp]
\caption{Examples of discovered events, where Time, Grid, QC\_cur, QC\_last, Top1 query word, Top1\_qc, and Description represents the start time and end time, grid coordinates, query counts in the current time period, query counts in the same period of last week, top 1 query word, top 1 query counts and the description of each event, respectively.}
\label{tab:events}
\begin{tabular}{lllllll}
\toprule
Time & Grid & QC\_cur & QC\_last & Top1 query word & Top1\_qc & Description  \\
\midrule
2017-04-08 14:00-20:00 &  (26, 39) & 3431 & 417 & Capital Gym & 2724 & Fish Leong Concert \\
2017-04-11 08:00-10:00 &  (24, 38) & 447 & 93 & Beijing Shangri-La Restaurant & 304 & IBM Data Scientist Forum \\
2017-04-15 08:00-16:00 &  (13, 47) & 4551 & 2202 & Beijing Botanical Garden & 3849 & Spring outing \\
2017-04-15 16:00-20:00 &  (21, 34) & 2173 & 207 & Letv sports center & 1831 & Chou Chuan-huing Concert \\
2017-04-30 08:00-18:00 &  (22, 47) & 7283 & 3607 & Summer Palace & 7149 & Summer Palace (May Day) \\
2017-04-30 08:00-18:00 &  (26, 46) & 3691 & 1582 & Tsinghua University & 3102 & 106th Anniversary of THU \\
\bottomrule
\end{tabular}
\end{table*}

\subsection{Query Sub-dataset}
\label{sec:query_sub_dataset}

\begin{figure}[htbp]
\centering
{\includegraphics[width=0.45\textwidth]{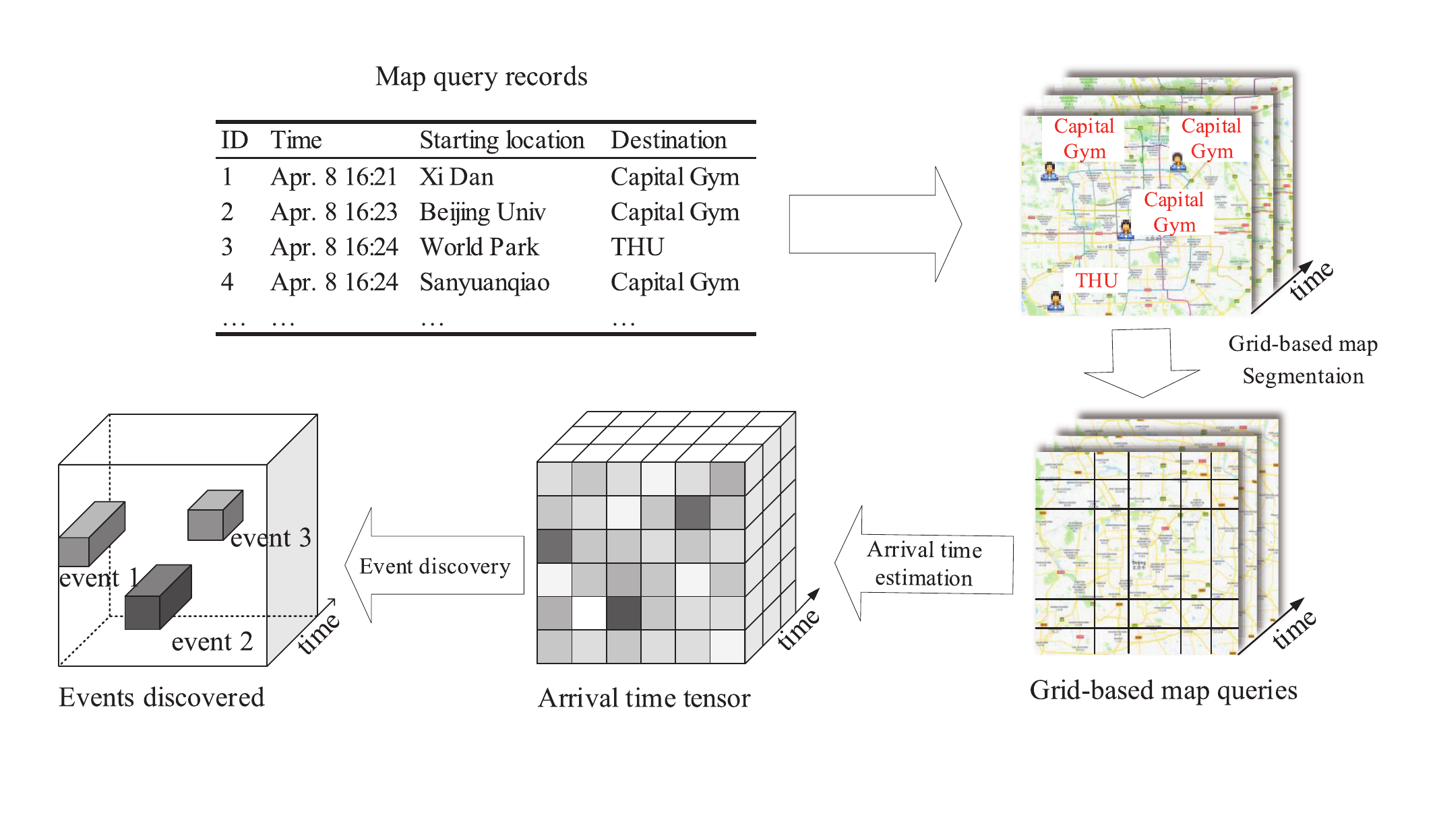}}
\caption{The flowchart of the mining of potential traffic in queries. A set of map queries is segmented into grids, then we can estimate the arrival time at each query's destination, thus constructing an arrival time tensor. An event discovery algorithm is used to discover the events from the arrival time tensor.}
\label{fig:querypreprocess}
\end{figure}

This sub-dataset was collected in Beijing, China between April 1, 2017 and May 31, 2017, from the Baidu Map\footnote{https://map.baidu.com} application. Two modes of map queries are provided by the Baidu Map: one is called ``location search'', which includes the searches of a specific place; the other is called ``route search'', which recommends a navigation route from one location to another. This sub-dataset contains about 114 million user queries, each of which records the user ID  (anonymised), search time-stamp, coordinate of the current location, coordinate and query word of starting location  (in ``route search''), coordinate and query word of destination. Note that if the query mode is ``location search'', the starting location is the same as the current location. The top/left of the Figure \ref{fig:querypreprocess} shows several records of the query sub-dataset. This dataset is pre-processed as follows, with the statistics of the dataset after processing given in Table \ref{tab:querystats}.

\begin{itemize}
\item To eliminate redundancy, only the last query will be retained if a single user submitted several queries in 10 minutes.
\item  The queries whose current locations are 2km (or more) away from the starting locations will be eliminated due to the assumption that the users are more likely to go to their searched destinations if they are currently close to the searched starting location.
\item Since the filtered starting locations are all within 2km from current locations, the starting time is estimated according to the distance between starting location and current location, with a speed of 3.6km/h (by walk).
\item As shown in Figure \ref{fig:querypreprocess}, the map is partitioned into a $R \times C$ grid map according to the lon/lat bounding box of  (116.10, 39.69, 116.71, 40.18), where $R=72, C=68$ and the width and height of a grid are both about 1km.
\item For each query, since the exact time when the user arrives at the destination is not available, an estimated arrival time $t_d^i$ for ``route search'' is calculated according to the query mode with a speed of 30km/h (by car), 20km/h (by bus), 10km/h (by bike) or 3.6km/h (by walk). And the estimated arrival time $t_d^i$ for ``location search'' is calculated according to the distance between the starting location and destination, with a speed of 20km/h (by bus, distance $>$ 2km) or 3.6km/h (by walk, distance $\leq$ 2km).
\end{itemize}

A query $q^i$ can be represented as $q^i= (t_s^i, s^i, d^i, x_s^i, y_s^i, x_d^i, y_d^i)$ where $i=1, 2, \dots, n$. We can calculate $(x_s^i, y_s^i, x_d^i, y_d^i)$ based on the grid map, where $t_s^i$, $t_d^i$, $s^i$, $d^i$, $(x_s^i, y_s^i)$, $(x_d^i, y_d^i)$, $n$ stand for the starting time-stamp, estimated arrival time, query word of the starting location, query word of the destination, coordinates of the starting location and the destination, and the number of all queries, respectively. 

\begin{table}[htbp]
\centering
\caption{Statistics of the query sub-dataset in Beijing between April 1 and May 31, 2017.}
\label{tab:querystats}
\begin{tabular}{ll}
\toprule
Items & \#  \\
\midrule
Filtered queries & 114,658,750 \\
Query words & 17,210,732 \\
Average distance/query & 12km \\
Average travel time/query & 46 minutes \\
\bottomrule
\end{tabular}
\end{table}

We note that a user may not go to the destination that they have searched, however, they will be much more likely to go to this destination if it relates to a public event that is occurring. To account for this, we declare an ``event" (see Sec.\ref{eventdiscovery}) to have occurred when query counts for a place are much higher than usual over a short period of time. For example, assume that we had received a lot of queries for Capital Gym as a destination and the arrival time was around 6 PM on April 8, 2017. As the number of queries is much more than usual we postulate that there could be an ``event" here (in this case the ``Fish Leong Concert"). The users who submitted these queries have a high probability of going to the Capital Gym for the concert. For all queries $\bm{Q}=\{q^i|i = 1, 2, \dots, n\}$, we can construct arrival time tensor $\bm{D}=\{d_{x, y, t}\}$, where $x=1, 2, \dots, C$, $y=1, 2, \dots, R$, $t=1, 2, \dots, T$, and $T=5,856$ ($61\text{-day} \times 24\text{-hour} \times 4\text{-quarter}$) is the total time-stamps (since we aggregate the queries every 15 minutes). We define $d_{x, y, t}$ as:
\begin{equation}
	\label{eq:sdxyt}
    d_{x, y, t} = |\{q^k|x_d^k=x, y_d^k=y, t_d^k=t\}|
\end{equation}
\noindent where $|\cdot|$ denotes the cardinality of a set. We will use the arrival time tensor $\bm{D}$ to discover the events in Sec.\ref{eventdiscovery} .

\subsubsection{Event Discovery}
\label{eventdiscovery}
The density $\rho_d = \frac{N}{R \times C\times T}$ of arrival time tensor $\bm{D}$ is 4, which implies $\bm{D}$ is very sparse. We will find the spatiotemporal ranges whose query counts are much more than usual, using the following definitions.

\begin{definition}[Moment]
    \label{def:moment}
    A tuple $m =  (x, y, t)$ is a moment if 
    \begin{eqnarray}
	\label{eq:moment}
    && d_{x, y, t-\Delta t} > 0 \\
    && d_{x, y, t} - d_{x, y, t-\Delta t} > \zeta \\ 
    && \frac{d_{x, y, t} - d_{x, y, t-\Delta t}}{d_{x, y, t-\Delta t}} > \eta
  	\end{eqnarray}    
\end{definition}
We denote $\mathcal{M}$ a set of all moments.

\begin{definition}[Event]
    \label{def:event}
    A tuple $E =  (x, y, t_s, t_d)$ is an event if 
    \begin{eqnarray}
    \label{eq:event}
    && t_d - t_s > \epsilon \\
    && \forall t \in [t_s, t_d], \quad m =  (x, y, t) \in \mathcal{M} \\
    && m =  (x, y, t_s - 1) \not\in \mathcal{M} \land m =  (x, y, t_d + 1) \not\in \mathcal{M}
    \end{eqnarray}
\end{definition}

According to the Def.\ref{def:moment} and Def.\ref{def:event}, let $\eta = 0.2$, $\zeta = 300$, $\Delta t = 672$  ($7 \times 24 \times 4$, a week), $\epsilon = 4$ (an hour), 932 events are discovered. The number of all the event queries and time steps is 2,336,114 and 15,892, respectively. Thus the density $\rho_e$ of all the events is 147, which is much larger than $\rho_d=4$. Table \ref{tab:events} shows some events that we discovered. Several kinds of events are presented, including concerts, forums, places of interest and anniversaries. For each event, one can find that the query counts are much more than that in last week, and the top 1 query word is highly related to the event. Not only that, more than 80\% of the query counts comes from the top 1 query word.

\subsection{Traffic Speed Sub-dataset}

\begin{figure}[htbp]
\centering
\includegraphics[width=0.45\textwidth]{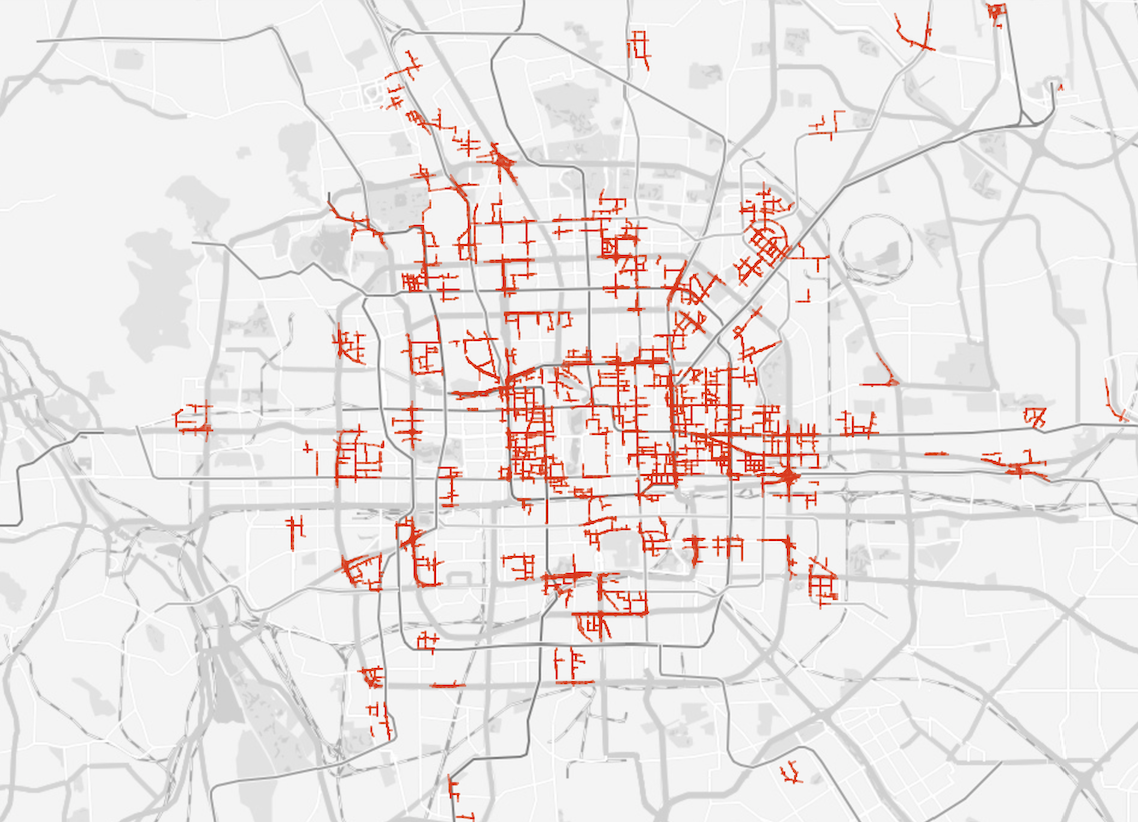}
\caption{Spatial distribution of the road segments in Beijing}
\label{fig:roadnetwork}
\end{figure}

We also collect the traffic speed data for the same area and during the same time period as the query sub-dataset. The origin traffic speed dataset contains the traffic speed of $\sim$450k road segments. What we are interested in are the traffic conditions of those road segments which are close to the events. Thus we collect a sub-dataset whose road segments are nearby the destinations where events held. This sub-dataset contains 15,073 road segments covering approximately 738.91 km. Table \ref{tab:trafficstats} and Figure  \ref{fig:roadnetwork} shows the statistics and spatial distribution of these road segments, respectively. They are all in the 6th ring road, which is the most crowded area of Beijing. The traffic speed of each road segment is recorded per minute. Since the traffic speed sub-dataset is from real-world urban areas, the traffic lights would have a great impact on the traffic speed, leading to the traffic speed varies greatly. For instance, the traffic speed may differ 20km/h between two consecutive minutes. To make the traffic speed predictable, for each road segment, we use simple moving average \footnote{https://en.wikipedia.org/wiki/Moving\_average} with a 15-minute time window to smooth the traffic speed sub-dataset and sample the traffic speed per 15 minutes. 

\begin{table}[htbp]
\centering
\caption{Statistics of the traffic speed sub-dataset}
\label{tab:trafficstats}
\begin{tabular}{ll}
\toprule
Items & \#  \\
\midrule
Road segments & 15,073 \\
Total length & 738.91 km \\
Interval & 15 minutes \\
Time & April 1, 2017 - May 31, 2017 \\
Total records & 265,967,808 \\
lon/lat bounding box &  (116.10, 39.69, 116.71, 40.18) \\
\bottomrule
\end{tabular}
\end{table}

\begin{table}[htbp]
\centering
\caption{Examples of geographical attributes of each road segment.}
\label{tab:linkproperties}
\begin{tabular}{lll}
\toprule
Field & Type & Description  \\
\midrule
link\_id & Char (13) & road segment id \\
width & Char (3) & width, 15: <=3.0m; 30:  (3.0m, 5.0m); \\
 & & 55:  (5.5m, 13m); 130: >13m \\
direction & Char (1) & direction, 0: unknown, default two-way; \\
 & & 1: two-way; 2: single-way, from start \\
 & & node to end node; 3: single-way, from \\
 & & end node to start node \\
snodeid & Char (13) & start node id \\
enodeid & Char (13) & end node id \\
snodegps & Char (30) & gps coordinate (lon, lat) of start node\\
enodegps & Char (30) & gps coordinate (lon, lat) of end node \\
length & Char (8) & length (kilo-meter) \\
speedclass & Char (1) & speed limit (km/h), 1: >130; 2:  (100, 130); \\
 & & 3:  (90, 100); 4:  (70, 90); 5:  (50, 70); \\
 & & 6:  (30, 50);  7:  (11, 30); 8: <11 \\
lanenum & Char (1) & number of lanes, 1: 1; 2: 2 or 3; 3: >=4 \\
\bottomrule
\end{tabular}
\end{table}

\begin{table*}[htbp]
\caption{Comparison of different datasets for traffic speed prediction.}
\label{tab:cmpdatasets}

\begin{tabular}{lccccccc}
\toprule
Datasets & Scale & Road info. & Road net. & Auxiliary info. & Highway & Urban & Available\tabularnewline
\midrule
Subset of PeMS &  &  & \multirow{5}{*}{$\surd$} &  & \multirow{5}{*}{$\surd$} &  & \multirow{5}{*}{$\surd$}\tabularnewline
State Route 22, Garden Grove \cite{yang2017ensemble} & 9 &  &  &  &  &  & \tabularnewline
PeMSD7 (S) \cite{yu2017spatio} & 228 &  &  &  &  &  & \tabularnewline
San Francisco Bay area \cite{he2013improving} & 943 &  &  &  &  &  & \tabularnewline
PeMSD7 (L) \cite{yu2017spatio} & 1,026 &  &  &  &  &  & \tabularnewline
\midrule
Subset of Beijing &  &  & \multirow{7}{*}{$\surd$} &  &  &  & \tabularnewline
Ring road around Beijing \cite{ma2015long} & 2 &  &  &  &  & $\surd$ & \tabularnewline
Beijing 4th ring road \cite{tang2017improved}  & 3 &  &  &  &  & $\surd$ & \tabularnewline
Beijing 2nd/3rd ring road  \cite{wang2016traffic} & 80 &  &  &  & $\surd$ &  & \tabularnewline
Beijing 2nd/3rd ring road \cite{wang2016traffic} & 122 &  &  &  & $\surd$ &  & \tabularnewline
Bejing taxi dataset \cite{ma2017learning} & 236 &  &  &  & $\surd$ & $\surd$ & \tabularnewline
Bejing taxi dataset \cite{ma2017learning} & 352 &  &  &  & $\surd$ & $\surd$ & \tabularnewline
\midrule
I-80 in California \cite{duan2016starima} & 6 &  & $\surd$ &  & $\surd$ &  & $\surd$\tabularnewline
Busan Metropolitan City \cite{kim2016comparison} & 10 &  & $\surd$ &  &  & $\surd$ & \tabularnewline
California PATH \cite{bezuglov2016short} & 12 &  &  &  & $\surd$ &  & \tabularnewline
Corridor in Orlando \cite{qi2014hidden} & 71 &  &  &  & $\surd$ &  & \tabularnewline
Rome dataset \cite{fusco2016short} & 120 &  & $\surd$ &  &  & $\surd$ & \tabularnewline
D100 \cite{gulaccar2016short} & 122 &  &  & weather &  & $\surd$ & \tabularnewline
Bedok area \cite{dauwels2014predicting} & 226 &  & $\surd$ &  & $\surd$ & $\surd$ & \tabularnewline
Los Angeles \cite{deng2016latent} & 1,642 &  & $\surd$ &  & $\surd$ & $\surd$ & \tabularnewline
Los Angeles \cite{deng2016latent} & 4,048 &  & $\surd$ &  & $\surd$ & $\surd$ & \tabularnewline
Dallas-Forth Worth area \cite{hasanzadeh2017graph} & 4,764 &  &  &  & $\surd$ &  & \tabularnewline
Subnetwork in Singapore \cite{asif2014spatiotemporal} & 5,024 &  & $\surd$ &  & $\surd$ & $\surd$ & \tabularnewline
\midrule
\textbf{Q-Traffic Dataset} & 15,073 & $\surd$ & $\surd$ & map query & $\surd$ & $\surd$ & $\surd$\tabularnewline
\toprule
\end{tabular}

\end{table*}

\subsection{Road Network Sub-dataset}

Due to the spatiotemporal dependencies of traffic data, the topology of the road network would help to predict traffic. Table \ref{tab:linkproperties} shows the fields of the road network sub-dataset. For each road segment in the traffic speed sub-dataset, the road network sub-dataset provides the starting node (\textit{snode}) and ending node (\textit{enode}) of the road segment, based on which the topology of the road network can be built. In addition, the sub-dataset also provides various geographical attributes of each road segment, such as width, length, speed limit and the number of lanes. Furthermore, we also provide the social attributes such as weekdays, weekends, public holidays, peak hours and off-peak hours.

\subsection{Comparison with Other Datasets}

Table \ref{tab:cmpdatasets} shows the comparison of different datasets for traffic speed prediction. The most popular dataset for traffic prediction is Caltrans Performance Measurement System (PeMS)\footnote{http://pems.dot.ca.gov/}. However, it doesn't provide road attributes and other auxiliary information. In the past few years, researchers have performed experiments with small or (and) private datasets. The release of Q-Traffic, a large-scale public available dataset with offline (geographical and social attributes, road network) and online (crowd map queries) information, should lead to an improvement of the research of traffic prediction.

\section{Methodologies}
\label{methods}
In this section, we will introduce the definition of traffic speed prediction problem and the hybrid model to integrate three auxiliary domains one by one.

\subsection{Problem Definition}
In a mere temporal model, the prediction of traffic speed can be formalised as forecasting future traffic speed $\{\bm{v}_{t+1}, \bm{v}_{t+2}, \dots, \bm{v}_{t+t'}\}$ given previous traffic speed $\{\bm{v}_1, \bm{v}_2, \dots, \bm{v}_t\}$. The Q-Traffic dataset makes it possible to combine benefits from multiple auxiliary domains including spatial relation within local road network, offline geographical and social attributes and online crowd search queries. 

\begin{figure}[htbp]
\centering
{\includegraphics[width=0.45\textwidth]{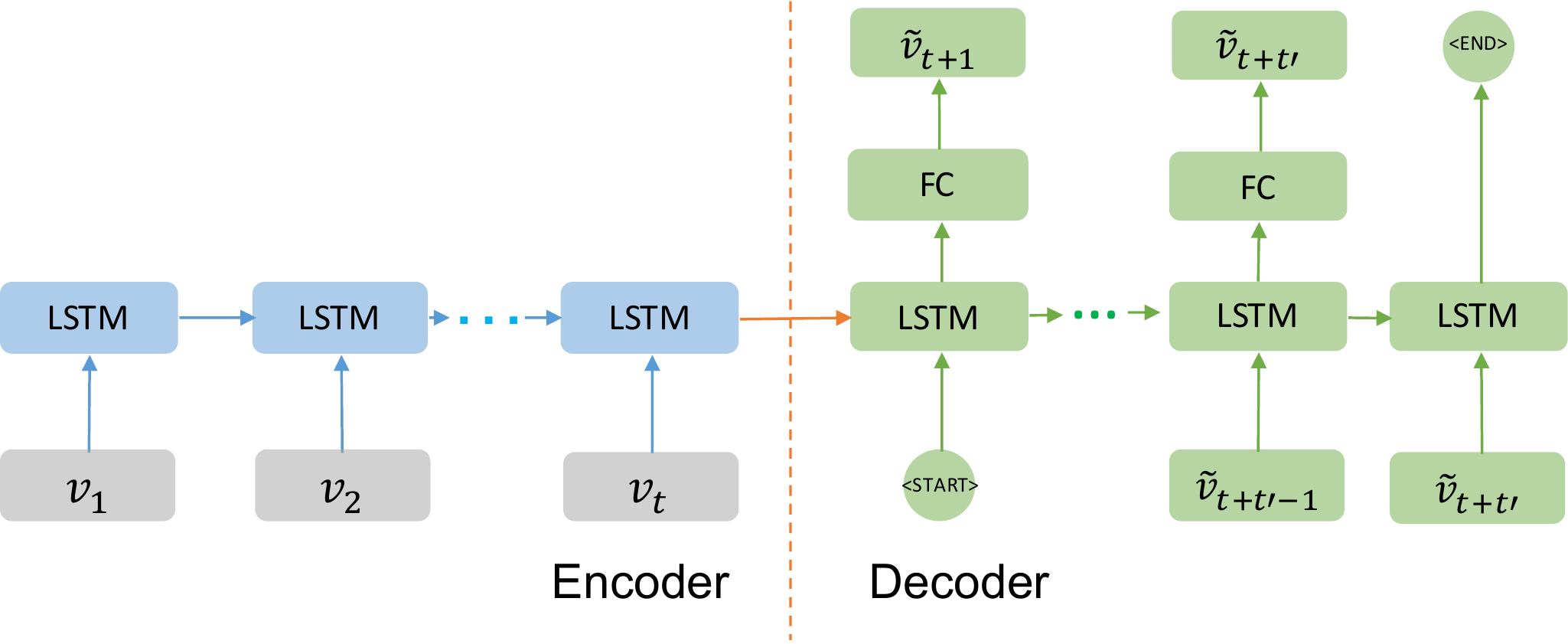}}
\caption{Seq2Seq: The Sequence to Sequence model predicts future traffic speed $\{ \tilde{\bm{v}}_{t+1}, \tilde{\bm{v}}_{t+2}, \dots, \tilde{\bm{v}}_{t+t'} \}$, given the previous traffic speed $\{ \bm{v}_{1}, \bm{v}_{2}, ... \bm{v}_{t} \}$. } 
\label{fig:model_seq2seq}
\end{figure}

\subsection{Seq2Seq}

The fundamental network to be applied on the Q-Traffic for mere temporal patterns is the Sequence to Sequence (Seq2Seq) model \cite{cho2014learning,sutskever2014sequence}. Both the encoder and decoder can be constructed based on LSTM \cite{hochreiter1997long}. The Seq2Seq model with LSTM has achieved a great success on different tasks such as speech recognition  \cite{graves2014towards}, machine translation  \cite{sutskever2014sequence} and video question answering  \cite{venugopalan2015sequence}. 

Figure \ref{fig:model_seq2seq} shows the architecture of the Seq2Seq model for traffic prediction. The encoder embeds the input traffic speed sequence $\{\bm{v}_{1}, \bm{v}_{2}, \dots, \bm{v}_{t}\}$ and the final hidden state of the encoder is fed into the decoder, which learns to predict the future traffic speed $\{\tilde{\bm{v}}_{t+1}, \tilde{\bm{v}}_{t+2}, \dots, \tilde{\bm{v}}_{t+t'}\}$. Hybrid model that integrates the auxiliary information will be proposed based on the Seq2Seq model.

\begin{figure}[htbp]
\centering
\includegraphics[width=0.45\textwidth]{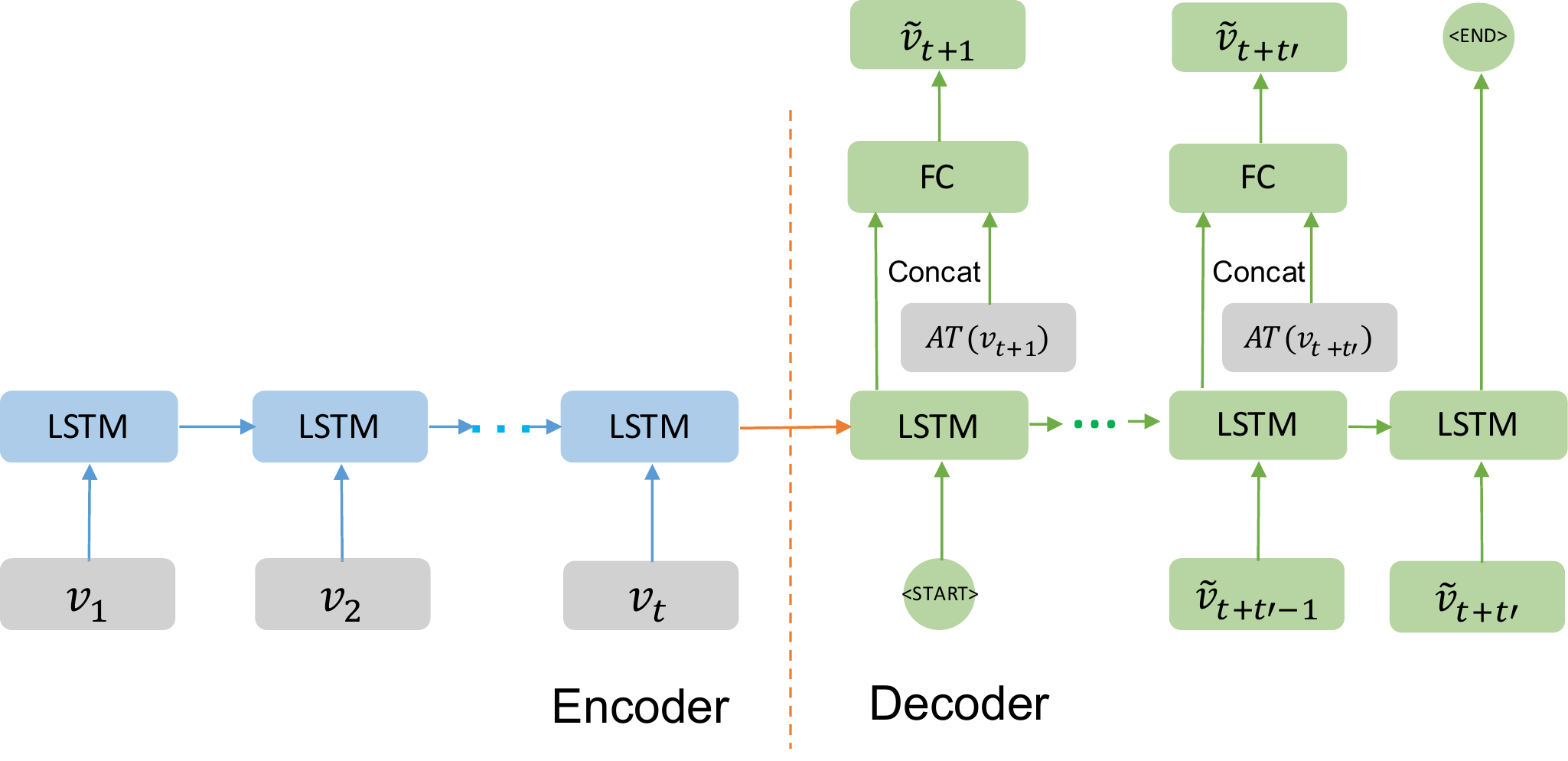}
\caption{Seq2Seq + Attributes: The model incorporates offline geographical and social attributes $\{\bm{AT} (\bm{v}_{t+1}), \bm{AT} (\bm{v}_{t+2}), \dots, \bm{AT} (\bm{v}_{t+t'}) \}$ into the decoder of the Seq2Seq model.}
\label{fig:model_widedeep}
\end{figure}

\subsection{Seq2Seq + Attributes}

As aforementioned, it is beneficial to introduce both geographical and social attributes into traffic sequence learning instead of merely utilising speed information. 

There are two kinds of offline attributes that have been extracted for wide and deep learning. 1) Geographical attributes including width, direction, speed limit and the number of lanes of each road segment (Table \ref{tab:linkproperties}). 2) Social attributes including information on public holidays, workdays, peak hours and off-peak hours.  

Motivated by \cite{cheng2016wide}, we utilise a wide and deep network that combine the benefits from deep learning and feature engineering. These attributes $\bm{AT} (\bm{v}_t)$ are concatenated directly into the decoder network of the Seq2Seq model and the encoder network remains identical as shown in Figure  \ref{fig:model_widedeep}.

\begin{figure}[htbp]
\centering
\includegraphics[width=0.45\textwidth]{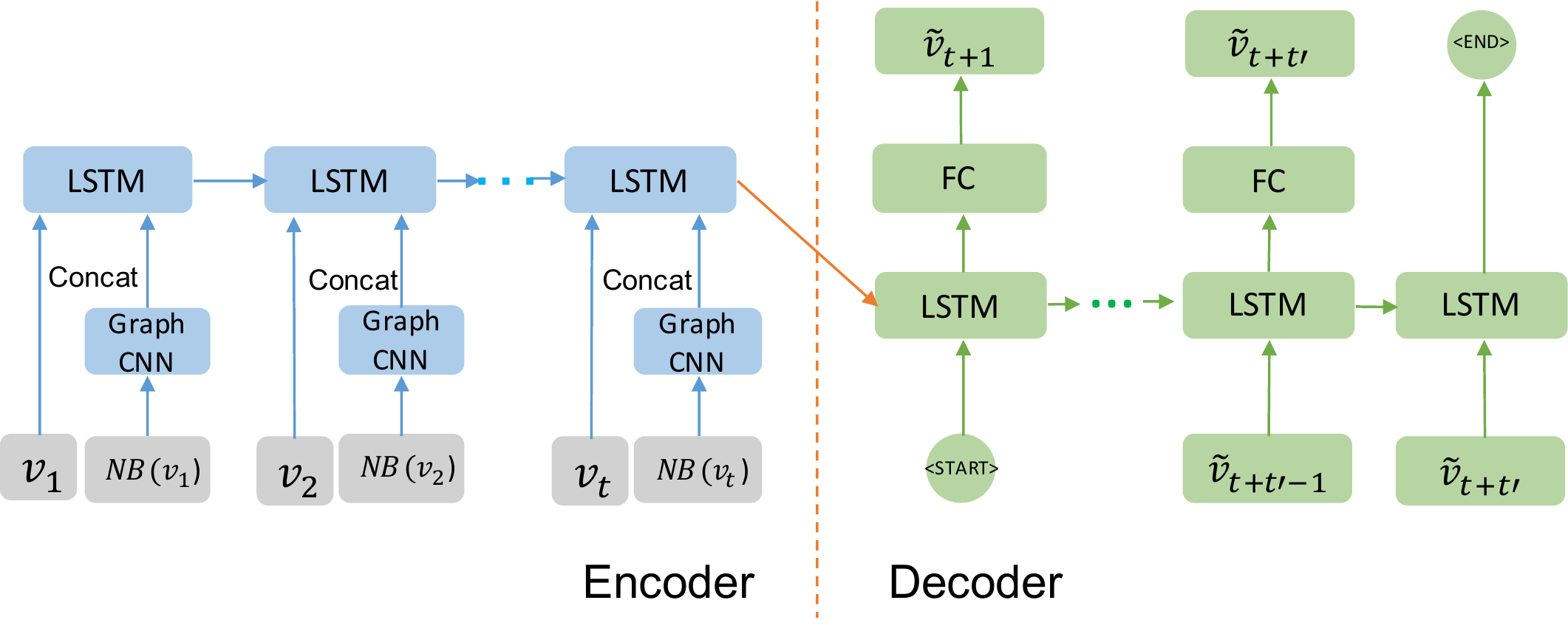}
\caption{Seq2Seq + Spatial Relation: Traffic speed of neighbouring road segments $\{\bm{NB} (\bm{v}_1), \bm{NB} (\bm{v}_2), \dots, \bm{NB} (\bm{v}_t)\}$ is embedded into the encoder of Seq2Seq model by using Graph Convolutional Neural Network. The neighbouring road segements are selected based on PageRank score, which reflects impact of neighbours on the central road segment.}
\label{fig:model_spatial}
\end{figure}

\subsection{Seq2Seq + Spatial Relation}

The traffic at a specific road segment can be affected by its neighbouring road segments  \cite{yu2017spatio}. However, the Seq2Seq model is designed to learn temporal dependencies but not spatial dependencies within road intersections. 

Graph Convolutional Neural Networks \cite{niepert2016learning} are applied to embed traffic of neighbouring road segments, $\bm{NB} (\bm{v}_t)$, into the encoder of the Seq2Seq model as shown in Figure \ref{fig:model_spatial}. As each road segment has a direction of traffic flow, the local road intersections can be constructed as a directed graph. Given a central road segment, five predecessors and five successors in the directed graph are selected based on PageRank score \cite{page1999pagerank}, which provides the relative spatial impact of neighbouring road segments on the central road segment.

\subsection{Seq2Seq + Query Impact}

\textbf{Correlation Analysis}. The query counts have a clear correlation with the traffic speed. For each event, we compute the average traffic speed of all the road segments within a range of 1km and measure the correlation between the average traffic speed and the corresponding query counts with a resolution of 15 minutes. As both of the variables are non-linear, Spearman's rank correlation coefficient is applied and the result $\rho=-0.52$ with a $P$-value$=1.23 \times 10^{-4}$ indicates a strong negative correlation between the average traffic speed and the query counts, making the prediction of traffic with the query sub-dataset promising.

\textbf{Query Impact}. The query impact $\bm{QI}$ measures the influence of queries on road segments. It is calculated based on the query counts and the spatial region that the query will influence. 

Given an event at location $A$, note that the queries come from different places $\{S_1, S_2, ...\}$, each query has different impact on different road segments around the location $A$. Thus, the query impact should consider the spatial locations of each query. As described in Section \ref{sec:query_sub_dataset}, let $Q=\{q^i|i = 1, 2, \dots, n\}$ denotes all the queries, where $q^i= (t_s^i, t_d^i, s^i, d^i, x_s^i, y_s^i, x_d^i, y_d^i)$. Road segments are denoted by $RS=\{l^i | i=1, 2, \dots, k\}$. The query impact $\bm{QI (l, t)}$ of queries on each road segments $l$ can be calculated by Algorithm \ref{algo:queryfe}. In the Algorithm \ref{algo:queryfe}, the function $dist$ in line 10 calculates the Euclidean distance of a point and a segment. The $h$ in line 11 is a decreasing function whose input is the distance $d_l$ and output is the impact $h (d_l)$ that query $q^i$ makes on the road segment $l$ at time $t_d^i$. For simplicity, we choose the exponential function $h (x)=exp (-\frac{x}{\sigma})$, where $\sigma$ is the impact factor.

\begin{algorithm}
\DontPrintSemicolon 
\KwIn{A set $Q=\{q^1, q^2, \dots, q^n\}$ of queries, where $q^i= (t_s^i, t_d^i, s^i, d^i, x_s^i, y_s^i, x_d^i, y_d^i)$, a set $RS=\{l^1, l^2, \dots, l^k\}$ of links, total time stamp $T$}
\KwOut{The query impact QI}
\textbf{Initialisation:} $QI (l, t) \gets 0$ \;
\For{$i \gets 1$ \textbf{to} $n$} {
  // return the longitude and latitude of a location \;
  $lonlat_s \gets lonlat (s^i)$ \;
  $lonlat_d \gets lonlat (d^i)$\;
  $seg_i \gets segment (lonlat_s, lonlat_d)$\;
  // return the set of road segments within 1km
  $L \gets nearroad (lonlat_d)$\;
  \For{each $l \in L$}{
    $lonlat_l \gets lonlat (l)$\;
    $d_l \gets dist (lonlat_{l}, seg_i)$\;
    $QI (l,t_d^i) \gets QI (l,t_d^i) + h (d_l)$
  }
}
\Return{$\bm{QI}$}\;
\caption{{\sc QueryImpact} Calculate the query impact}
\label{algo:queryfe}
\end{algorithm}

\begin{figure}[htbp]
\centering
\includegraphics[width=0.45\textwidth]{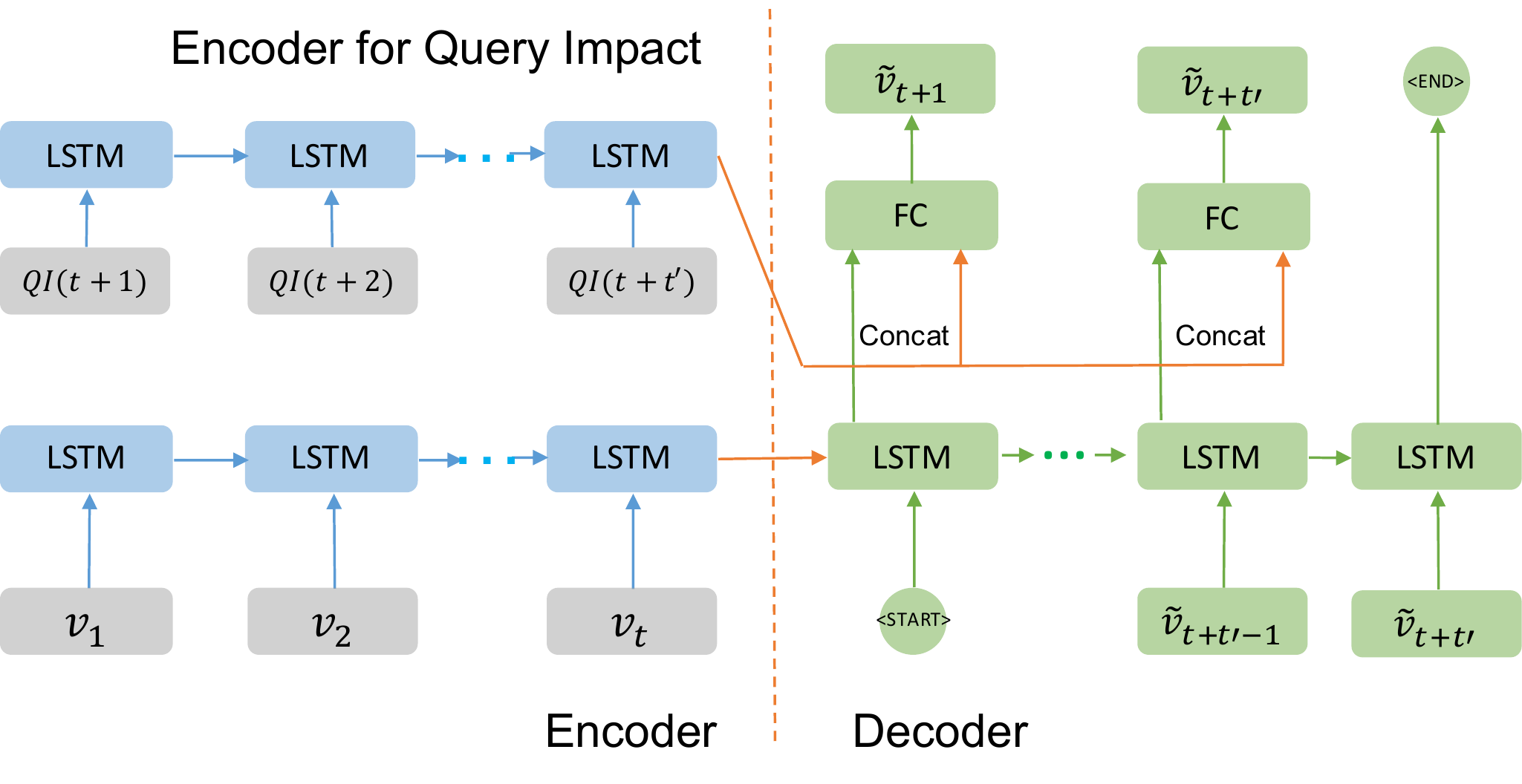}
\caption{Seq2Seq + Query Impact: The query impact $\bm{QI}$ are calculated based on the query counts and the spatial region that the queries can influence. An RNN encoder with LSTM is applied to encode the QI sequence. For simplicity, note that the $\bm{Q (t)}$ refers to $\bm{Q (l, t)}$ on corresponding road segment.}
\label{fig:model_query}
\end{figure}

The query impact $\{\bm{QI (l, t+1)}, \bm{QI (l, t+2)}, ..., \bm{QI (l, t+t')} \}$ are encoded by a RNN with LSTM and the final hidden state is concatenated into the decoder of the Seq2Seq model and the encoder remains identical as shown in Figure  \ref{fig:model_query}.

\begin{figure}[htbp]
\centering
\includegraphics[width=0.45\textwidth]{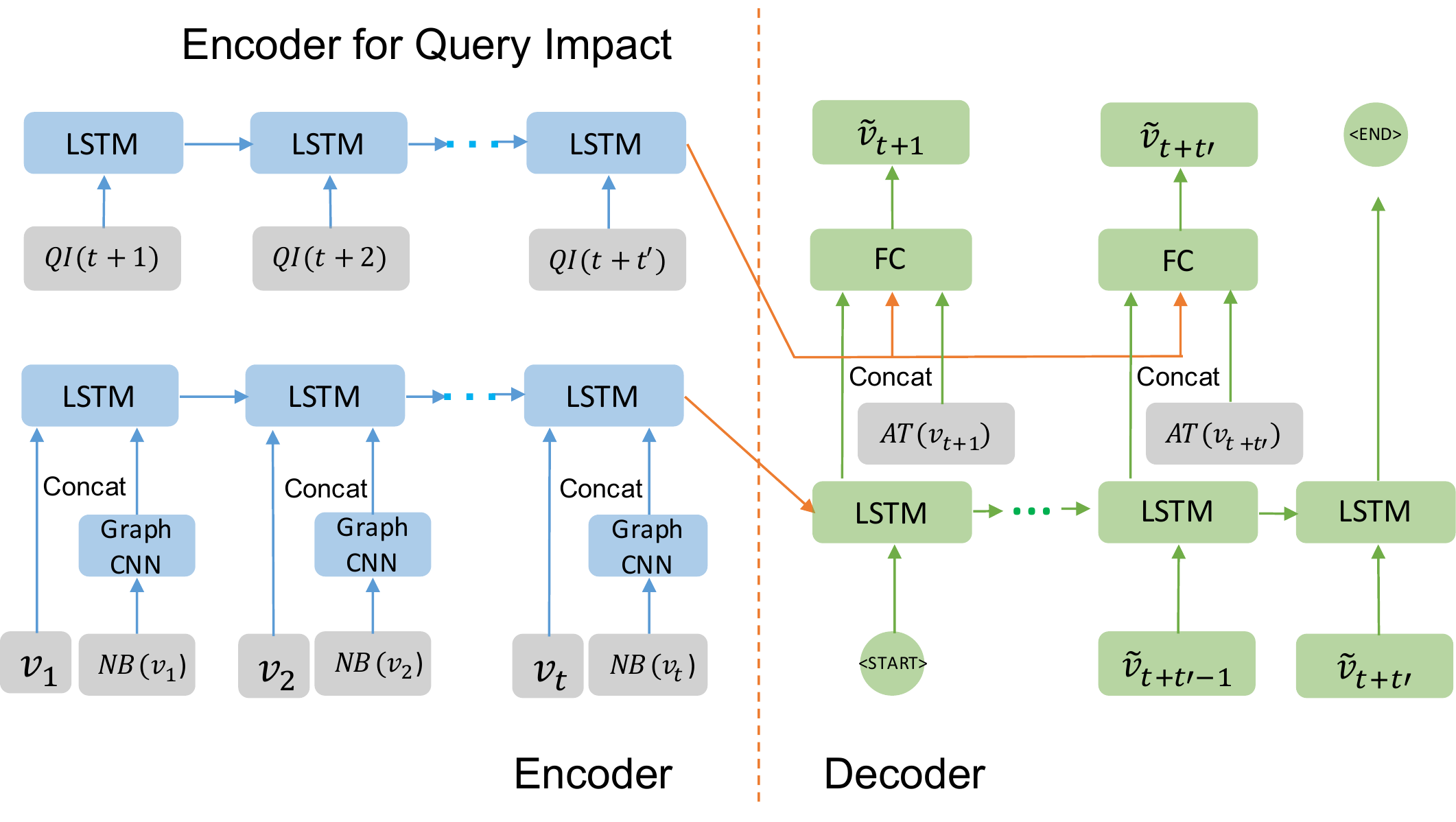}
\caption{Hybrid model integrates all three auxiliary domains including attributes $\bm{AT} (\bm{v}_t)$, spatial relation $\bm{NB} (\bm{v}_t)$, and query impact $\bm{QI(l, t)}$.}
\label{fig:model_hybrid}
\end{figure}

\subsection{Hybrid Model}

The hybrid model combines benefits from all available information domains. Figure  \ref{fig:model_hybrid} shows that the hybrid model enhances the Seq2Seq model with the attributes, the spatial relation, and the query impact. The spatial relation from $\bm{NB} (\bm{v}_t)$ are incorporated into the encoder of the Seq2Seq model, while the attributes $\bm{AT} (\bm{v}_t)$ and the query impact $\bm{QI(l, t)} $ are embedded into the decoder.

\section{Experiments}
\label{experiments}
In this section, we describe the compared methods, implementation details and evaluation metrics. We also discuss the results of different models.
\subsection{Compared methods}
\label{comparedmethods}

We compare our proposed model with the following methods.
\begin{itemize}
\item Random forests (RF) \cite{leshem2007traffic}: RF is a traditional machine learning method for regression, often used for traffic prediction;
\item Support vector regression  (SVR) \cite{jin2007simultaneously}: SVR is a version of SVM for regression, widely used in traffic prediction;
\item Deep neural networks including the Seq2Seq model, the Seq2Seq + Attributes  (AT) model, the Seq2Seq + Spatial relation  (NB) model, the Seq2Seq + Query Impact  (QI) model and the Hybrid Model that combines all three auxiliary domains.
\end{itemize}

\begin{table*}
\caption{$Err_T$ (\%): MAPE on the whole testing set. The results with the best performance are marked in bold.}
\label{tab:cmpall}
\begin{tabular}{llllllllll}
\toprule
Prediction & 15-min & 30-min & 45-min & 60-min & 75-min & 90-min & 105-min & 120-min & Overall\\
\midrule
RF &  6.00 & 9.15 & 10.20 & 10.66 & 10.98 & 11.21 & 11.39 & 11.56 & 10.14 \\
SVR  &  5.44 & 9.20 & 10.07 & 10.34 & 10.51 & 10.65 & 10.76 & 10.83 & 9.73 \\
\midrule
Seq2Seq  & 4.61 & 8.22 & 9.28 & 9.72 & 9.98 & 10.27 & 10.48 & 10.61 & 9.23\\
Seq2Seq+AT    & 4.53 & 8.06 & 9.09 & 9.48 & 9.70 & 9.84 & 9.93 & 10.01 & 8.83\\
Seq2Seq+NB & \textbf{4.52} & 8.05 & 9.07 & 9.45 & 9.67 & 9.83 & 9.93 & 9.99 & 8.81\\
\midrule
Seq2Seq+QI   & 4.58 & 8.01 & 8.95 & 9.31 & 9.51 & 9.66 & 9.80 & 9.94 & 8.72\\
Hybrid & \textbf{4.52} & \textbf{7.93} & \textbf{8.89} & \textbf{9.24 }& \textbf{9.43} & \textbf{9.56} & \textbf{9.69} & \textbf{9.78} & \textbf{8.63}\\
\bottomrule
\end{tabular}
\end{table*}

\begin{table*}
\caption{$Err_E$ (\%): MAPE during events on the testing set. The results with the best performance are marked in bold.}
\label{tab:cmpevent}
\begin{tabular}{llllllllll}
\toprule
Prediction & 15-min & 30-min & 45-min & 60-min & 75-min & 90-min & 105-min & 120-min & Overall\\
\midrule
RF & 6.14 & 9.51 & 10.81 & 11.45 & 11.84 & 12.13 & 12.38 & 12.56 & 10.85 \\
SVR & 5.64 & 9.56 & 10.59 & 11.02 & 11.32 & 11.56 & 11.73 & 11.83 & 10.41 \\
\midrule
Seq2Seq         & 4.76 & 8.52 & 9.87 & 10.52 & 10.91 & 11.31 & 11.60 & 11.80 & 9.91\\
Seq2Seq+AT    & 4.65 & 8.32 & 9.63 & 10.23 & 10.58 & 10.81 & 10.98 & 11.13 & 9.54\\
Seq2Seq+NB & 4.63 & 8.25 & 9.53 & 10.10 & 10.45 & 10.70 & 10.89 & 11.02 & 9.45\\
\midrule
Seq2Seq+QI   & 4.69 & 8.18 & 9.37 & 9.93  & 10.28 & 10.55 & 10.77 & 10.98 & 9.34\\
Hybrid & \textbf{4.61} & \textbf{8.09} &\textbf{ 9.30} & \textbf{9.84} & \textbf{10.16} & \textbf{10.39} & \textbf{10.60} & \textbf{10.76} & \textbf{9.22}\\
\bottomrule
\end{tabular}
\end{table*}

\subsection{Implementation Details}
\label{details}

The traffic speed with 15-minute moving average is sampled at 15-minute intervals. One-day traffic speed sequence is used as inputs to predict future 2-hour traffic speed. Thus the length of the input sequence is $t=96$ and the length of the output sequence is $t'=8$.

Given the traffic speed sequence $\{\bm{v}_{1}, \bm{v}_{2}, \dots, \bm{v}_{t}\}$, note that the standard Random Forests and SVR can only predict the traffic speed $\bm{v}_{t+1}$, whose goals are slightly different from our Seq2Seq model. So, on the testing stage, we treat prior forecasts as observations and use them for subsequent forecasts. For the Random Forests, the number of trees and the maximum depth are both 10. For SVR, we set $C=1, \epsilon=0.1$ and use RBF kernel.  We use scikit-learn \cite{pedregosa2011scikit} to implement the Random Forests and SVR. 

All of the deep neural networks are implemented by TensorLayer  \cite{haoTL2017} which is a deep learning and reinforcement learning library based on TensorFlow  \cite{abadi2016tensorflow}. Stochastic gradient descent using the Adam optimiser  \cite{kingma2014adam} is applied to update trainable parameters with mini-batch size 128. The deep neural networks are trained on a single NVIDIA GeForce GTX TITAN X GPU with 12GB memory.

The dimension of LSTM hidden state is set as 128. The social attributes have 6 dimensions and the geographical attributes have 131 dimensions. Half of the data  (the first month) is used as the training set and the other half (the second month) is used for testing. The impact factor for calculating the query impact is $\sigma=150$.

\subsection{Evaluation metrics}
\label{metrics}
The mean absolute percentage error  (MAPE) is used to evaluate the performance for comparisons, which is defined as
\begin{eqnarray}
MAPE &=& \frac{1}{T} \sum_{t=1}^{T}{\arrowvert \frac{\bm{v}_t - \tilde{\bm{v}}_t}{\bm{v}_t} \arrowvert}
\end{eqnarray}
where $\bm{v}_t$ and $\tilde{\bm{v}}_t$ are the actual and predicted traffic speed at time $t$, respectively. Two error rates are calculated. 1) $Err_T$ computes the MAPE across the whole testing set. 2) $Err_E$ only computes the MAPE during events on the testing set.

\subsection{Results and discussion}
\label{results}

Table \ref{tab:cmpall} and Table \ref{tab:cmpevent} show quantitative comparison between different models. On the whole testing set, the hybrid model achieves the best performance with lowest overall MAPE 8.63\% and 2-hour forecasting MAPE 9.78\%. Compared with the common Seq2Seq model, auxiliary information (e.g., attributes, spatial relation and query impact) is able to improve the accuracy and the query impact is more effective than the other two domains.

It is more challenging to predict traffic when events happen, which is abnormally burst in traffic (Figure \ref{fig:querytraffic}). Table \ref{tab:cmpevent} shows MAPE during events and the prediction is less accurate than that during normal conditions, especially for forecasting traffic more than one hour. With query impact, the hybrid model achieves the best performance with overall MAPE 9.22\% and the Seq2Seq + Query Impact model achieves the second best accuracy with overall MAPE 9.34\%.

Therefore, with appropriate integration of three kinds of auxiliary information, the deep encoder-decoder sequence networks can boost the traffic speed prediction during both events period and whole time period. Compared with the other two auxiliary domains, the query impact information has demonstrated larger improvement and promising application in the practical scenario.

\section{Related Work}
\label{relatedwork}

\subsection{Traffic Prediction}
\label{trafficpred}
Traffic prediction is a critical component in ITS. The methods of traffic prediction can be classified into parametric and non-parametric methods. Autoregressive integrated moving average (ARIMA) \cite{ahmed1979analysis} is a widely used parametric technique for traffic prediction, which is based on the assumption that the traffic prediction is in a stationary process. However, ARIMA and its family require high computational resources and are not suitable for large-scale traffic data. Due to the stochastic and non-linear nature of traffic, researchers have paid much attention to the non-parametric methods, such as RF \cite{leshem2007traffic}, SVR \cite{jin2007simultaneously}, online support vector regression (OL-SVR) \cite{castro2009online}, Bayesian network \cite{sun2006bayesian} and neural networks like ANN \cite{vlahogianni2005optimized}. Recently, some deep learning methods are proposed for traffic prediction such as stacked autoencoders (SAEs) \cite{lv2015traffic}, deep belief network \cite{huang2014deep}. Most of them consider the traffic prediction for highways, whose traffic condition are relatively stable. However, in real-world cities, the traffic lights may have a great impact on the traffic speed, thus the traffic speed may vary greatly and previous methods may no longer work. \cite{ma2015long} proposed to use long short-term memory network (LSTM) to predict traffic speed in cities. However, they only choose two points to conduct their experiments, which may be insufficient for practical applications.

\subsection{Traffic Prediction via Auxiliary Domains}
A few researchers have attempted to predict the traffic with related multimodal data. \cite{he2013improving} \cite{liu2016collective} proposed an optimization framework to use traffic indicators extracted from social media with location information to model traffic sequence via linear regression. However, the traffic is a nonlinear system, so using a linear regression may be insufficient. \cite{ma2015long} proposed an LSTM to forecast traffic speed given microwave detector data. However, in real-world road systems, only a small fraction of the road segments are deployed with sensors. For those road segments without sensors, previous methods may no longer work. Spatial dependencies among nodes can also improve traffic prediction \cite{asif2014spatiotemporal} and  \cite{yu2017spatio,ma2017learning} propose models based on CNN in order to incorporate spatial information.

\section{Conclusions}
\label{conclusions}
As an important part of intelligent transportation systems, traffic prediction is limited by complex traffic environment in large cities and accessibility of high-quality multi-modal dataset. In this paper, we firstly introduce and release a large-scale traffic dataset, Q-Traffic, addressing the obstacle of the limited dataset. This dataset contains integrated online and offline auxiliary information including crowd map queries, road intersections and geographical and social attributes. Then we appropriately integrate traffic prediction with crowd map queries, road intersections and geographical and social attributes into a sequence to sequence learning framework, respectively. At last, a hybrid model is proposed to combine the benefits from all three auxiliary information domains and proved to be superior to compared methods, addressing the obstacles of traffic prediction with offline geographical factors and online potential influence. As for future work, with the release of the dataset, it is promising to attract more people to devise more approaches for accurate and real-time traffic prediction.

\begin{acks}
We would like to thank the support from Baidu Map and ZJU-Imperial Joint Lab. This work was supported by the National Basic Research Program (973) of China (No. 2015CB352302), the National Natural Science Foundation of China (Nos. 61625107, U1611461, U1509206), the Key Program of Zhejiang Province, China (No. 2015C01027), the Chinese Knowledge Center for Engineering Sciences and Technology, and the Fundamental Research Funds for the Central Universities, China. Jingqing Zhang was funded by LexisNexis HPCC Systems Academic Program.
\end{acks}

\bibliographystyle{ACM-Reference-Format}
\balance
\bibliography{traffic_pred_kdd}


\begin{thebibliography}{39}


\ifx \showCODEN    \undefined \def \showCODEN     #1{\unskip}     \fi
\ifx \showDOI      \undefined \def \showDOI       #1{#1}\fi
\ifx \showISBNx    \undefined \def \showISBNx     #1{\unskip}     \fi
\ifx \showISBNxiii \undefined \def \showISBNxiii  #1{\unskip}     \fi
\ifx \showISSN     \undefined \def \showISSN      #1{\unskip}     \fi
\ifx \showLCCN     \undefined \def \showLCCN      #1{\unskip}     \fi
\ifx \shownote     \undefined \def \shownote      #1{#1}          \fi
\ifx \showarticletitle \undefined \def \showarticletitle #1{#1}   \fi
\ifx \showURL      \undefined \def \showURL       {\relax}        \fi
\providecommand\bibfield[2]{#2}
\providecommand\bibinfo[2]{#2}
\providecommand\natexlab[1]{#1}
\providecommand\showeprint[2][]{arXiv:#2}

\bibitem[\protect\citeauthoryear{Abadi, Agarwal, Barham, Brevdo, Chen, Citro,
  Corrado, Davis, Dean, Devin, et~al\mbox{.}}{Abadi et~al\mbox{.}}{2016}]%
        {abadi2016tensorflow}
\bibfield{author}{\bibinfo{person}{Mart{\'\i}n Abadi}, \bibinfo{person}{Ashish
  Agarwal}, \bibinfo{person}{Paul Barham}, \bibinfo{person}{Eugene Brevdo},
  \bibinfo{person}{Zhifeng Chen}, \bibinfo{person}{Craig Citro},
  \bibinfo{person}{Greg~S Corrado}, \bibinfo{person}{Andy Davis},
  \bibinfo{person}{Jeffrey Dean}, \bibinfo{person}{Matthieu Devin},
  {et~al\mbox{.}}} \bibinfo{year}{2016}\natexlab{}.
\newblock \showarticletitle{Tensorflow: Large-scale machine learning on
  heterogeneous distributed systems}.
\newblock \bibinfo{journal}{\emph{arXiv preprint arXiv:1603.04467}}
  (\bibinfo{year}{2016}).
\newblock


\bibitem[\protect\citeauthoryear{Ahmed and Cook}{Ahmed and Cook}{1979}]%
        {ahmed1979analysis}
\bibfield{author}{\bibinfo{person}{Mohammed~S Ahmed} {and}
  \bibinfo{person}{Allen~R Cook}.} \bibinfo{year}{1979}\natexlab{}.
\newblock \bibinfo{booktitle}{\emph{Analysis of freeway traffic time-series
  data by using Box-Jenkins techniques}}.
\newblock Number 722.
\newblock


\bibitem[\protect\citeauthoryear{Asif, Dauwels, Goh, Oran, Fathi, Xu, Dhanya,
  Mitrovic, and Jaillet}{Asif et~al\mbox{.}}{2014}]%
        {asif2014spatiotemporal}
\bibfield{author}{\bibinfo{person}{Muhammad~Tayyab Asif},
  \bibinfo{person}{Justin Dauwels}, \bibinfo{person}{Chong~Yang Goh},
  \bibinfo{person}{Ali Oran}, \bibinfo{person}{Esmail Fathi},
  \bibinfo{person}{Muye Xu}, \bibinfo{person}{Menoth~Mohan Dhanya},
  \bibinfo{person}{Nikola Mitrovic}, {and} \bibinfo{person}{Patrick Jaillet}.}
  \bibinfo{year}{2014}\natexlab{}.
\newblock \showarticletitle{Spatiotemporal patterns in large-scale traffic
  speed prediction}.
\newblock \bibinfo{journal}{\emph{IEEE Transactions on Intelligent
  Transportation Systems}} \bibinfo{volume}{15}, \bibinfo{number}{2}
  (\bibinfo{year}{2014}), \bibinfo{pages}{794--804}.
\newblock


\bibitem[\protect\citeauthoryear{Bezuglov and Comert}{Bezuglov and
  Comert}{2016}]%
        {bezuglov2016short}
\bibfield{author}{\bibinfo{person}{Anton Bezuglov} {and}
  \bibinfo{person}{Gurcan Comert}.} \bibinfo{year}{2016}\natexlab{}.
\newblock \showarticletitle{Short-term freeway traffic parameter prediction:
  Application of grey system theory models}.
\newblock \bibinfo{journal}{\emph{Expert Systems with Applications}}
  \bibinfo{volume}{62} (\bibinfo{year}{2016}), \bibinfo{pages}{284--292}.
\newblock


\bibitem[\protect\citeauthoryear{Castro-Neto, Jeong, Jeong, and
  Han}{Castro-Neto et~al\mbox{.}}{2009}]%
        {castro2009online}
\bibfield{author}{\bibinfo{person}{Manoel Castro-Neto},
  \bibinfo{person}{Young-Seon Jeong}, \bibinfo{person}{Myong-Kee Jeong}, {and}
  \bibinfo{person}{Lee~D Han}.} \bibinfo{year}{2009}\natexlab{}.
\newblock \showarticletitle{Online-SVR for short-term traffic flow prediction
  under typical and atypical traffic conditions}.
\newblock \bibinfo{journal}{\emph{Expert systems with applications}}
  \bibinfo{volume}{36}, \bibinfo{number}{3} (\bibinfo{year}{2009}),
  \bibinfo{pages}{6164--6173}.
\newblock


\bibitem[\protect\citeauthoryear{Cheng, Koc, Harmsen, Shaked, Chandra, Aradhye,
  Anderson, Corrado, Chai, Ispir, et~al\mbox{.}}{Cheng et~al\mbox{.}}{2016}]%
        {cheng2016wide}
\bibfield{author}{\bibinfo{person}{Heng-Tze Cheng}, \bibinfo{person}{Levent
  Koc}, \bibinfo{person}{Jeremiah Harmsen}, \bibinfo{person}{Tal Shaked},
  \bibinfo{person}{Tushar Chandra}, \bibinfo{person}{Hrishi Aradhye},
  \bibinfo{person}{Glen Anderson}, \bibinfo{person}{Greg Corrado},
  \bibinfo{person}{Wei Chai}, \bibinfo{person}{Mustafa Ispir}, {et~al\mbox{.}}}
  \bibinfo{year}{2016}\natexlab{}.
\newblock \showarticletitle{Wide \& deep learning for recommender systems}. In
  \bibinfo{booktitle}{\emph{Proceedings of the 1st Workshop on Deep Learning
  for Recommender Systems}}. ACM, \bibinfo{pages}{7--10}.
\newblock


\bibitem[\protect\citeauthoryear{Cho, Van~Merri{\"e}nboer, Gulcehre, Bahdanau,
  Bougares, Schwenk, and Bengio}{Cho et~al\mbox{.}}{2014}]%
        {cho2014learning}
\bibfield{author}{\bibinfo{person}{Kyunghyun Cho}, \bibinfo{person}{Bart
  Van~Merri{\"e}nboer}, \bibinfo{person}{Caglar Gulcehre},
  \bibinfo{person}{Dzmitry Bahdanau}, \bibinfo{person}{Fethi Bougares},
  \bibinfo{person}{Holger Schwenk}, {and} \bibinfo{person}{Yoshua Bengio}.}
  \bibinfo{year}{2014}\natexlab{}.
\newblock \showarticletitle{Learning phrase representations using RNN
  encoder-decoder for statistical machine translation}.
\newblock \bibinfo{journal}{\emph{arXiv preprint arXiv:1406.1078}}
  (\bibinfo{year}{2014}).
\newblock


\bibitem[\protect\citeauthoryear{Dauwels, Aslam, Asif, Zhao, Vie, Cichocki, and
  Jaillet}{Dauwels et~al\mbox{.}}{2014}]%
        {dauwels2014predicting}
\bibfield{author}{\bibinfo{person}{Justin Dauwels}, \bibinfo{person}{Aamer
  Aslam}, \bibinfo{person}{Muhammad~Tayyab Asif}, \bibinfo{person}{Xinyue
  Zhao}, \bibinfo{person}{Nikola~Mitro Vie}, \bibinfo{person}{Andrzej
  Cichocki}, {and} \bibinfo{person}{Patrick Jaillet}.}
  \bibinfo{year}{2014}\natexlab{}.
\newblock \showarticletitle{Predicting traffic speed in urban transportation
  subnetworks for multiple horizons}. In \bibinfo{booktitle}{\emph{Control
  Automation Robotics \& Vision (ICARCV), 2014 13th International Conference
  on}}. IEEE, \bibinfo{pages}{547--552}.
\newblock


\bibitem[\protect\citeauthoryear{Deng, Shahabi, Demiryurek, Zhu, Yu, and
  Liu}{Deng et~al\mbox{.}}{2016}]%
        {deng2016latent}
\bibfield{author}{\bibinfo{person}{Dingxiong Deng}, \bibinfo{person}{Cyrus
  Shahabi}, \bibinfo{person}{Ugur Demiryurek}, \bibinfo{person}{Linhong Zhu},
  \bibinfo{person}{Rose Yu}, {and} \bibinfo{person}{Yan Liu}.}
  \bibinfo{year}{2016}\natexlab{}.
\newblock \showarticletitle{Latent space model for road networks to predict
  time-varying traffic}. In \bibinfo{booktitle}{\emph{Proceedings of the 22nd
  ACM SIGKDD International Conference on Knowledge Discovery and Data Mining}}.
  ACM, \bibinfo{pages}{1525--1534}.
\newblock


\bibitem[\protect\citeauthoryear{Dong, Supratak, Mai, Liu, Oehmichen, Yu, and
  Guo}{Dong et~al\mbox{.}}{2017}]%
        {haoTL2017}
\bibfield{author}{\bibinfo{person}{Hao Dong}, \bibinfo{person}{Akara Supratak},
  \bibinfo{person}{Luo Mai}, \bibinfo{person}{Fangde Liu},
  \bibinfo{person}{Axel Oehmichen}, \bibinfo{person}{Simiao Yu}, {and}
  \bibinfo{person}{Yike Guo}.} \bibinfo{year}{2017}\natexlab{}.
\newblock \showarticletitle{{TensorLayer: A Versatile Library for Efficient
  Deep Learning Development}}.
\newblock \bibinfo{journal}{\emph{ACM Multimedia}} (\bibinfo{year}{2017}).
\newblock
\urldef\tempurl%
\url{http://tensorlayer.org}
\showURL{%
\tempurl}


\bibitem[\protect\citeauthoryear{Duan, Mao, Zhang, and Wang}{Duan
  et~al\mbox{.}}{2016}]%
        {duan2016starima}
\bibfield{author}{\bibinfo{person}{Peibo Duan}, \bibinfo{person}{Guoqiang Mao},
  \bibinfo{person}{Changsheng Zhang}, {and} \bibinfo{person}{Shangbo Wang}.}
  \bibinfo{year}{2016}\natexlab{}.
\newblock \showarticletitle{STARIMA-based traffic prediction with time-varying
  lags}. In \bibinfo{booktitle}{\emph{Intelligent Transportation Systems
  (ITSC), 2016 IEEE 19th International Conference on}}. IEEE,
  \bibinfo{pages}{1610--1615}.
\newblock


\bibitem[\protect\citeauthoryear{Fusco, Colombaroni, and Isaenko}{Fusco
  et~al\mbox{.}}{2016}]%
        {fusco2016short}
\bibfield{author}{\bibinfo{person}{Gaetano Fusco}, \bibinfo{person}{Chiara
  Colombaroni}, {and} \bibinfo{person}{Natalia Isaenko}.}
  \bibinfo{year}{2016}\natexlab{}.
\newblock \showarticletitle{Short-term speed predictions exploiting big data on
  large urban road networks}.
\newblock \bibinfo{journal}{\emph{Transportation Research Part C: Emerging
  Technologies}}  \bibinfo{volume}{73} (\bibinfo{year}{2016}),
  \bibinfo{pages}{183--201}.
\newblock


\bibitem[\protect\citeauthoryear{Graves and Jaitly}{Graves and Jaitly}{2014}]%
        {graves2014towards}
\bibfield{author}{\bibinfo{person}{Alex Graves} {and} \bibinfo{person}{Navdeep
  Jaitly}.} \bibinfo{year}{2014}\natexlab{}.
\newblock \showarticletitle{Towards end-to-end speech recognition with
  recurrent neural networks}. In \bibinfo{booktitle}{\emph{Proceedings of the
  31st International Conference on Machine Learning (ICML-14)}}.
  \bibinfo{pages}{1764--1772}.
\newblock


\bibitem[\protect\citeauthoryear{G{\"u}la{\c{c}}ar, Yaslan, and
  Oktu{\u{g}}}{G{\"u}la{\c{c}}ar et~al\mbox{.}}{2016}]%
        {gulaccar2016short}
\bibfield{author}{\bibinfo{person}{Halil G{\"u}la{\c{c}}ar},
  \bibinfo{person}{Yusuf Yaslan}, {and} \bibinfo{person}{Sema~F Oktu{\u{g}}}.}
  \bibinfo{year}{2016}\natexlab{}.
\newblock \showarticletitle{Short term traffic speed prediction using different
  feature sets and sensor clusters}. In \bibinfo{booktitle}{\emph{Network
  Operations and Management Symposium (NOMS), 2016 IEEE/IFIP}}. IEEE,
  \bibinfo{pages}{1265--1268}.
\newblock


\bibitem[\protect\citeauthoryear{Hasanzadeh, Liu, Duffield, Narayanan, and
  Chigoy}{Hasanzadeh et~al\mbox{.}}{2017}]%
        {hasanzadeh2017graph}
\bibfield{author}{\bibinfo{person}{Arman Hasanzadeh}, \bibinfo{person}{Xi Liu},
  \bibinfo{person}{Nick Duffield}, \bibinfo{person}{Krishna~R Narayanan}, {and}
  \bibinfo{person}{Byron Chigoy}.} \bibinfo{year}{2017}\natexlab{}.
\newblock \showarticletitle{A Graph Signal Processing Approach For Real-Time
  Traffic Prediction In Transportation Networks}.
\newblock \bibinfo{journal}{\emph{arXiv preprint arXiv:1711.06954}}
  (\bibinfo{year}{2017}).
\newblock


\bibitem[\protect\citeauthoryear{He, Shen, Divakaruni, Wynter, and Lawrence}{He
  et~al\mbox{.}}{2013}]%
        {he2013improving}
\bibfield{author}{\bibinfo{person}{Jingrui He}, \bibinfo{person}{Wei Shen},
  \bibinfo{person}{Phani Divakaruni}, \bibinfo{person}{Laura Wynter}, {and}
  \bibinfo{person}{Rick Lawrence}.} \bibinfo{year}{2013}\natexlab{}.
\newblock \showarticletitle{Improving Traffic Prediction with Tweet
  Semantics.}. In \bibinfo{booktitle}{\emph{IJCAI}}.
  \bibinfo{pages}{1387--1393}.
\newblock


\bibitem[\protect\citeauthoryear{Hochreiter and Schmidhuber}{Hochreiter and
  Schmidhuber}{1997}]%
        {hochreiter1997long}
\bibfield{author}{\bibinfo{person}{Sepp Hochreiter} {and}
  \bibinfo{person}{J{\"u}rgen Schmidhuber}.} \bibinfo{year}{1997}\natexlab{}.
\newblock \showarticletitle{Long short-term memory}.
\newblock \bibinfo{journal}{\emph{Neural computation}} \bibinfo{volume}{9},
  \bibinfo{number}{8} (\bibinfo{year}{1997}), \bibinfo{pages}{1735--1780}.
\newblock


\bibitem[\protect\citeauthoryear{Huang, Song, Hong, and Xie}{Huang
  et~al\mbox{.}}{2014}]%
        {huang2014deep}
\bibfield{author}{\bibinfo{person}{Wenhao Huang}, \bibinfo{person}{Guojie
  Song}, \bibinfo{person}{Haikun Hong}, {and} \bibinfo{person}{Kunqing Xie}.}
  \bibinfo{year}{2014}\natexlab{}.
\newblock \showarticletitle{Deep architecture for traffic flow prediction: deep
  belief networks with multitask learning}.
\newblock \bibinfo{journal}{\emph{IEEE Transactions on Intelligent
  Transportation Systems}} \bibinfo{volume}{15}, \bibinfo{number}{5}
  (\bibinfo{year}{2014}), \bibinfo{pages}{2191--2201}.
\newblock


\bibitem[\protect\citeauthoryear{Jin, Zhang, and Yao}{Jin
  et~al\mbox{.}}{2007}]%
        {jin2007simultaneously}
\bibfield{author}{\bibinfo{person}{Xuexiang Jin}, \bibinfo{person}{Yi Zhang},
  {and} \bibinfo{person}{Danya Yao}.} \bibinfo{year}{2007}\natexlab{}.
\newblock \showarticletitle{Simultaneously prediction of network traffic flow
  based on PCA-SVR}.
\newblock \bibinfo{journal}{\emph{Advances in Neural Networks--ISNN 2007}}
  (\bibinfo{year}{2007}), \bibinfo{pages}{1022--1031}.
\newblock


\bibitem[\protect\citeauthoryear{Kim, Kim, and Ryu}{Kim et~al\mbox{.}}{2016}]%
        {kim2016comparison}
\bibfield{author}{\bibinfo{person}{Seyoung Kim}, \bibinfo{person}{Jeongmin
  Kim}, {and} \bibinfo{person}{Kwang~Ryel Ryu}.}
  \bibinfo{year}{2016}\natexlab{}.
\newblock \showarticletitle{Comparison of Different k-NN Models for Speed
  Prediction in an Urban Traffic Network}.
\newblock \bibinfo{journal}{\emph{World Academy of Science, Engineering and
  Technology, International Journal of Computer and Information Engineering}}
  \bibinfo{volume}{3}, \bibinfo{number}{2} (\bibinfo{year}{2016}).
\newblock


\bibitem[\protect\citeauthoryear{Kingma and Ba}{Kingma and Ba}{2014}]%
        {kingma2014adam}
\bibfield{author}{\bibinfo{person}{Diederik Kingma} {and}
  \bibinfo{person}{Jimmy Ba}.} \bibinfo{year}{2014}\natexlab{}.
\newblock \showarticletitle{Adam: A method for stochastic optimization}.
\newblock \bibinfo{journal}{\emph{arXiv preprint arXiv:1412.6980}}
  (\bibinfo{year}{2014}).
\newblock


\bibitem[\protect\citeauthoryear{Leshem and Ritov}{Leshem and Ritov}{2007}]%
        {leshem2007traffic}
\bibfield{author}{\bibinfo{person}{Guy Leshem} {and} \bibinfo{person}{Yaacov
  Ritov}.} \bibinfo{year}{2007}\natexlab{}.
\newblock \showarticletitle{Traffic flow prediction using adaboost algorithm
  with random forests as a weak learner}. In
  \bibinfo{booktitle}{\emph{Proceedings of World Academy of Science,
  Engineering and Technology}}, Vol.~\bibinfo{volume}{19}.
  \bibinfo{pages}{193--198}.
\newblock


\bibitem[\protect\citeauthoryear{Liu, Kong, and Li}{Liu et~al\mbox{.}}{2016}]%
        {liu2016collective}
\bibfield{author}{\bibinfo{person}{Xinyue Liu}, \bibinfo{person}{Xiangnan
  Kong}, {and} \bibinfo{person}{Yanhua Li}.} \bibinfo{year}{2016}\natexlab{}.
\newblock \showarticletitle{Collective Traffic Prediction with Partially
  Observed Traffic History using Location-Based Social Media}. In
  \bibinfo{booktitle}{\emph{Proceedings of the 25th ACM International on
  Conference on Information and Knowledge Management}}. ACM,
  \bibinfo{pages}{2179--2184}.
\newblock


\bibitem[\protect\citeauthoryear{Lv, Duan, Kang, Li, and Wang}{Lv
  et~al\mbox{.}}{2015}]%
        {lv2015traffic}
\bibfield{author}{\bibinfo{person}{Yisheng Lv}, \bibinfo{person}{Yanjie Duan},
  \bibinfo{person}{Wenwen Kang}, \bibinfo{person}{Zhengxi Li}, {and}
  \bibinfo{person}{Fei-Yue Wang}.} \bibinfo{year}{2015}\natexlab{}.
\newblock \showarticletitle{Traffic flow prediction with big data: a deep
  learning approach}.
\newblock \bibinfo{journal}{\emph{IEEE Transactions on Intelligent
  Transportation Systems}} \bibinfo{volume}{16}, \bibinfo{number}{2}
  (\bibinfo{year}{2015}), \bibinfo{pages}{865--873}.
\newblock


\bibitem[\protect\citeauthoryear{Ma, Dai, He, Ma, Wang, and Wang}{Ma
  et~al\mbox{.}}{2017}]%
        {ma2017learning}
\bibfield{author}{\bibinfo{person}{Xiaolei Ma}, \bibinfo{person}{Zhuang Dai},
  \bibinfo{person}{Zhengbing He}, \bibinfo{person}{Jihui Ma},
  \bibinfo{person}{Yong Wang}, {and} \bibinfo{person}{Yunpeng Wang}.}
  \bibinfo{year}{2017}\natexlab{}.
\newblock \showarticletitle{Learning traffic as images: a deep convolutional
  neural network for large-scale transportation network speed prediction}.
\newblock \bibinfo{journal}{\emph{Sensors}} \bibinfo{volume}{17},
  \bibinfo{number}{4} (\bibinfo{year}{2017}), \bibinfo{pages}{818}.
\newblock


\bibitem[\protect\citeauthoryear{Ma, Tao, Wang, Yu, and Wang}{Ma
  et~al\mbox{.}}{2015}]%
        {ma2015long}
\bibfield{author}{\bibinfo{person}{Xiaolei Ma}, \bibinfo{person}{Zhimin Tao},
  \bibinfo{person}{Yinhai Wang}, \bibinfo{person}{Haiyang Yu}, {and}
  \bibinfo{person}{Yunpeng Wang}.} \bibinfo{year}{2015}\natexlab{}.
\newblock \showarticletitle{Long short-term memory neural network for traffic
  speed prediction using remote microwave sensor data}.
\newblock \bibinfo{journal}{\emph{Transportation Research Part C: Emerging
  Technologies}}  \bibinfo{volume}{54} (\bibinfo{year}{2015}),
  \bibinfo{pages}{187--197}.
\newblock


\bibitem[\protect\citeauthoryear{Niepert, Ahmed, and Kutzkov}{Niepert
  et~al\mbox{.}}{2016}]%
        {niepert2016learning}
\bibfield{author}{\bibinfo{person}{Mathias Niepert}, \bibinfo{person}{Mohamed
  Ahmed}, {and} \bibinfo{person}{Konstantin Kutzkov}.}
  \bibinfo{year}{2016}\natexlab{}.
\newblock \showarticletitle{Learning convolutional neural networks for graphs}.
  In \bibinfo{booktitle}{\emph{International conference on machine learning}}.
  \bibinfo{pages}{2014--2023}.
\newblock


\bibitem[\protect\citeauthoryear{Page, Brin, Motwani, and Winograd}{Page
  et~al\mbox{.}}{1999}]%
        {page1999pagerank}
\bibfield{author}{\bibinfo{person}{Lawrence Page}, \bibinfo{person}{Sergey
  Brin}, \bibinfo{person}{Rajeev Motwani}, {and} \bibinfo{person}{Terry
  Winograd}.} \bibinfo{year}{1999}\natexlab{}.
\newblock \bibinfo{booktitle}{\emph{The PageRank citation ranking: Bringing
  order to the web.}}
\newblock \bibinfo{type}{{T}echnical {R}eport}. \bibinfo{institution}{Stanford
  InfoLab}.
\newblock


\bibitem[\protect\citeauthoryear{Pedregosa, Varoquaux, Gramfort, Michel,
  Thirion, Grisel, Blondel, Prettenhofer, Weiss, Dubourg,
  et~al\mbox{.}}{Pedregosa et~al\mbox{.}}{2011}]%
        {pedregosa2011scikit}
\bibfield{author}{\bibinfo{person}{Fabian Pedregosa}, \bibinfo{person}{Ga{\"e}l
  Varoquaux}, \bibinfo{person}{Alexandre Gramfort}, \bibinfo{person}{Vincent
  Michel}, \bibinfo{person}{Bertrand Thirion}, \bibinfo{person}{Olivier
  Grisel}, \bibinfo{person}{Mathieu Blondel}, \bibinfo{person}{Peter
  Prettenhofer}, \bibinfo{person}{Ron Weiss}, \bibinfo{person}{Vincent
  Dubourg}, {et~al\mbox{.}}} \bibinfo{year}{2011}\natexlab{}.
\newblock \showarticletitle{Scikit-learn: Machine learning in Python}.
\newblock \bibinfo{journal}{\emph{Journal of Machine Learning Research}}
  \bibinfo{volume}{12}, \bibinfo{number}{Oct} (\bibinfo{year}{2011}),
  \bibinfo{pages}{2825--2830}.
\newblock


\bibitem[\protect\citeauthoryear{Qi and Ishak}{Qi and Ishak}{2014}]%
        {qi2014hidden}
\bibfield{author}{\bibinfo{person}{Yan Qi} {and} \bibinfo{person}{Sherif
  Ishak}.} \bibinfo{year}{2014}\natexlab{}.
\newblock \showarticletitle{A Hidden Markov Model for short term prediction of
  traffic conditions on freeways}.
\newblock \bibinfo{journal}{\emph{Transportation Research Part C: Emerging
  Technologies}}  \bibinfo{volume}{43} (\bibinfo{year}{2014}),
  \bibinfo{pages}{95--111}.
\newblock


\bibitem[\protect\citeauthoryear{Sun, Zhang, and Yu}{Sun et~al\mbox{.}}{2006}]%
        {sun2006bayesian}
\bibfield{author}{\bibinfo{person}{Shiliang Sun}, \bibinfo{person}{Changshui
  Zhang}, {and} \bibinfo{person}{Guoqiang Yu}.}
  \bibinfo{year}{2006}\natexlab{}.
\newblock \showarticletitle{A Bayesian network approach to traffic flow
  forecasting}.
\newblock \bibinfo{journal}{\emph{IEEE Transactions on intelligent
  transportation systems}} \bibinfo{volume}{7}, \bibinfo{number}{1}
  (\bibinfo{year}{2006}), \bibinfo{pages}{124--132}.
\newblock


\bibitem[\protect\citeauthoryear{Sutskever, Vinyals, and Le}{Sutskever
  et~al\mbox{.}}{2014}]%
        {sutskever2014sequence}
\bibfield{author}{\bibinfo{person}{Ilya Sutskever}, \bibinfo{person}{Oriol
  Vinyals}, {and} \bibinfo{person}{Quoc~V Le}.}
  \bibinfo{year}{2014}\natexlab{}.
\newblock \showarticletitle{Sequence to sequence learning with neural
  networks}. In \bibinfo{booktitle}{\emph{Advances in neural information
  processing systems}}. \bibinfo{pages}{3104--3112}.
\newblock


\bibitem[\protect\citeauthoryear{Tang, Liu, Zou, Zhang, and Wang}{Tang
  et~al\mbox{.}}{2017}]%
        {tang2017improved}
\bibfield{author}{\bibinfo{person}{Jinjun Tang}, \bibinfo{person}{Fang Liu},
  \bibinfo{person}{Yajie Zou}, \bibinfo{person}{Weibin Zhang}, {and}
  \bibinfo{person}{Yinhai Wang}.} \bibinfo{year}{2017}\natexlab{}.
\newblock \showarticletitle{An improved fuzzy neural network for traffic speed
  prediction considering periodic characteristic}.
\newblock \bibinfo{journal}{\emph{IEEE Transactions on Intelligent
  Transportation Systems}} \bibinfo{volume}{18}, \bibinfo{number}{9}
  (\bibinfo{year}{2017}), \bibinfo{pages}{2340--2350}.
\newblock


\bibitem[\protect\citeauthoryear{Venugopalan, Rohrbach, Donahue, Mooney,
  Darrell, and Saenko}{Venugopalan et~al\mbox{.}}{2015}]%
        {venugopalan2015sequence}
\bibfield{author}{\bibinfo{person}{Subhashini Venugopalan},
  \bibinfo{person}{Marcus Rohrbach}, \bibinfo{person}{Jeffrey Donahue},
  \bibinfo{person}{Raymond Mooney}, \bibinfo{person}{Trevor Darrell}, {and}
  \bibinfo{person}{Kate Saenko}.} \bibinfo{year}{2015}\natexlab{}.
\newblock \showarticletitle{Sequence to sequence-video to text}. In
  \bibinfo{booktitle}{\emph{Proceedings of the IEEE international conference on
  computer vision}}. \bibinfo{pages}{4534--4542}.
\newblock


\bibitem[\protect\citeauthoryear{Vlahogianni, Karlaftis, and
  Golias}{Vlahogianni et~al\mbox{.}}{2005}]%
        {vlahogianni2005optimized}
\bibfield{author}{\bibinfo{person}{Eleni~I Vlahogianni},
  \bibinfo{person}{Matthew~G Karlaftis}, {and} \bibinfo{person}{John~C
  Golias}.} \bibinfo{year}{2005}\natexlab{}.
\newblock \showarticletitle{Optimized and meta-optimized neural networks for
  short-term traffic flow prediction: a genetic approach}.
\newblock \bibinfo{journal}{\emph{Transportation Research Part C: Emerging
  Technologies}} \bibinfo{volume}{13}, \bibinfo{number}{3}
  (\bibinfo{year}{2005}), \bibinfo{pages}{211--234}.
\newblock


\bibitem[\protect\citeauthoryear{Wang, Gu, Wu, Liu, and Xiong}{Wang
  et~al\mbox{.}}{2016}]%
        {wang2016traffic}
\bibfield{author}{\bibinfo{person}{Jingyuan Wang}, \bibinfo{person}{Qian Gu},
  \bibinfo{person}{Junjie Wu}, \bibinfo{person}{Guannan Liu}, {and}
  \bibinfo{person}{Zhang Xiong}.} \bibinfo{year}{2016}\natexlab{}.
\newblock \showarticletitle{Traffic speed prediction and congestion source
  exploration: A deep learning method}. In \bibinfo{booktitle}{\emph{Data
  Mining (ICDM), 2016 IEEE 16th International Conference on}}. IEEE,
  \bibinfo{pages}{499--508}.
\newblock


\bibitem[\protect\citeauthoryear{Yang, Wu, Du, He, and Chen}{Yang
  et~al\mbox{.}}{2017}]%
        {yang2017ensemble}
\bibfield{author}{\bibinfo{person}{Senyan Yang}, \bibinfo{person}{Jianping Wu},
  \bibinfo{person}{Yiman Du}, \bibinfo{person}{Yingqi He}, {and}
  \bibinfo{person}{Xu Chen}.} \bibinfo{year}{2017}\natexlab{}.
\newblock \showarticletitle{Ensemble Learning for Short-Term Traffic Prediction
  Based on Gradient Boosting Machine}.
\newblock \bibinfo{journal}{\emph{Journal of Sensors}}  \bibinfo{volume}{2017}
  (\bibinfo{year}{2017}).
\newblock


\bibitem[\protect\citeauthoryear{Yu, Yin, and Zhu}{Yu et~al\mbox{.}}{2017}]%
        {yu2017spatio}
\bibfield{author}{\bibinfo{person}{Bing Yu}, \bibinfo{person}{Haoteng Yin},
  {and} \bibinfo{person}{Zhanxing Zhu}.} \bibinfo{year}{2017}\natexlab{}.
\newblock \showarticletitle{Spatio-temporal Graph Convolutional Neural Network:
  A Deep Learning Framework for Traffic Forecasting}.
\newblock \bibinfo{journal}{\emph{arXiv preprint arXiv:1709.04875}}
  (\bibinfo{year}{2017}).
\newblock


\bibitem[\protect\citeauthoryear{Zhang, Wang, Wang, Lin, Xu, and Chen}{Zhang
  et~al\mbox{.}}{2011}]%
        {zhang2011data}
\bibfield{author}{\bibinfo{person}{Junping Zhang}, \bibinfo{person}{Fei-Yue
  Wang}, \bibinfo{person}{Kunfeng Wang}, \bibinfo{person}{Wei-Hua Lin},
  \bibinfo{person}{Xin Xu}, {and} \bibinfo{person}{Cheng Chen}.}
  \bibinfo{year}{2011}\natexlab{}.
\newblock \showarticletitle{Data-driven intelligent transportation systems: A
  survey}.
\newblock \bibinfo{journal}{\emph{IEEE Transactions on Intelligent
  Transportation Systems}} \bibinfo{volume}{12}, \bibinfo{number}{4}
  (\bibinfo{year}{2011}), \bibinfo{pages}{1624--1639}.
\newblock


\end{thebibliography}

\end{document}